\documentclass[lettersize,journal]{IEEEtran}
\usepackage{amsmath,amsfonts}
\usepackage{amssymb}
\usepackage{booktabs}
\usepackage{rotating}
\usepackage{multirow}
\usepackage{cuted}
\usepackage{capt-of}
\usepackage{placeins}
\usepackage{algorithmic}
\usepackage{algorithm}
\usepackage{array}
\usepackage[caption=false,font=normalsize,labelfont=sf,textfont=sf]{subfig}
\usepackage{textcomp}
\usepackage{stfloats}
\usepackage{url}
\usepackage{verbatim}
\usepackage{graphicx}
\usepackage{cite}
\usepackage{verbatim}
\usepackage{bm}
\usepackage[table]{xcolor}
\usepackage{bbding}
\usepackage[pagebackref,breaklinks,colorlinks]{hyperref}

\newcommand{\red}[1]{\textcolor{red}{ #1}}
\begin{document}

\title{UniRS: Unifying Multi-temporal Remote Sensing Tasks through Vision Language Models}

\author{Yujie Li,~\IEEEmembership{} Wenjia Xu$^{\ast}$,~\IEEEmembership{} Guangzuo Li,~\IEEEmembership{} Zijian Yu,~\IEEEmembership{}Zhiwei Wei,~\IEEEmembership{}Jiuniu Wang,~\IEEEmembership{}Mugen Peng,~\IEEEmembership{Fellow,~IEEE,}
        % <-this % stops a space
\thanks{Y. Li, W. Xu, Z. Yu and M. Peng are affiliated with the State Key Laboratory of Networking and Switching Technology, Beijing University of Posts and Telecommunications, Beijing 100876, China. G. Li is with the Aerospace Information Research Institute, Chinese Academy of Sciences, Beijing 100190, China. Z. Wei is affiliated with the School of Geographic Sciences, Hunan Normal University. J. Wang is with City University of Hong Kong, Hong Kong SAR.}% <-this % stops a space
\thanks{Corresponding author: W. Xu, xuwenjia@bupt.edu.cn}}

\markboth{IEEE Transactions on Geoscience and Remote Sensing}%
{UniRS: Unifying Multi-temporal Remote Sensing Tasks through Vision Language Models}

% 查一下这个部位的作用

\maketitle

\begin{abstract}
% Vision Language Models (VLMs) have recently shown promising prospects in remote sensing. 
% Numerous studies have proposed large-scale instruction-response datasets, leading to the development of domain-specific VLMs through fine-tuning. These models demonstrate impressive performance across various multimodal tasks in remote sensing. 
% However, previous works often require training dedicated models for each task or type of visual input (e.g., video), which consumes significant resources and falls short of the ideal interaction style we envision with AI assistants. To address this gap, we propose UniRS, the first VLM that integrates multimodal conversational tasks for the single image, dual-time image pair, and video remote sensing analysis within a single model. UniRS can perform single-image understanding tasks and seamlessly incorporate dual-time image pair and video understanding capabilities without altering the model's structure. Experiments demonstrate that UniRS achieves superior performance across tasks involving the three types of visual inputs (i.e., visual question answering, change captioning, and video classification). 
The domain gap between remote sensing imagery and natural images has recently received widespread attention and Vision-Language Models (VLMs) have demonstrated excellent generalization performance in remote sensing multimodal tasks. 
% However, current research is still in its early stages with insufficient exploration of the capabilities of VLMs, which is limited by a narrow range of remote sensing visual input types. 
However, current research is still limited in exploring how remote sensing VLMs handle different types of visual inputs.
To bridge this gap, we introduce \textbf{UniRS}, the first vision-language model \textbf{uni}fying multi-temporal \textbf{r}emote \textbf{s}ensing tasks across various types of visual input. UniRS supports single images, dual-time image pairs, and videos as input, enabling comprehensive remote sensing temporal analysis within a unified framework. We adopt a unified visual representation approach, enabling the model to accept various visual inputs. For dual-time image pair tasks, we customize a change extraction module to further enhance the extraction of spatiotemporal features. Additionally, we design a prompt augmentation mechanism tailored to the model's reasoning process, utilizing the prior knowledge of the general-purpose VLM to provide clues for UniRS. To promote multi-task knowledge sharing, the model is jointly fine-tuned on a mixed dataset. Experimental results show that UniRS achieves state-of-the-art performance across diverse tasks, including visual question answering, change captioning, and video scene classification, highlighting its versatility and effectiveness in unifying these multi-temporal remote sensing tasks. Our code and dataset will be released soon.
\end{abstract}


\section{Introduction}
% \IEEEPARstart{L}{arge} Vision-Language Models (VLMs)~\cite{liu2024visual} have recently opened up new possibilities for interaction between humans and machines. By inputting visual information along with corresponding instructions, VLMs can analyze the multimodal content and provide reasonable responses. This capability has attracted the attention of researchers in the remote sensing field. Researchers have developed large-scale remote sensing image instruction-following datasets~\cite{kuckreja2024geochat,zhan2024skyeyegpt,luo2024skysensegpt} and fine-tuned general vision-language conversational models~\cite{kuckreja2024geochat, zhan2024skyeyegpt, luo2024skysensegpt, pang2024h2rsvlm, muhtar2024lhrs} to enable these models understand images and videos in the remote sensing domain.
\IEEEPARstart{B}{y} aligning visual representations with textual feature space, Vision-Language Models (VLMs)~\cite{liu2024visual, alayrac2022flamingo, lin2024vila, chen2023minigpt, li2022blip, li2023blip} have brought Large Language Models (LLMs)~\cite{touvron2023llama, ouyang2022training, bai2023qwen, brown2020language} visual image understanding and multimodal instruction following capabilities. Leveraging the extensive prior knowledge and powerful multi-task reasoning capabilities of LLMs, general-purpose VLMs exhibit excellent generalization and robustness across various vision-language multimodal tasks after fine-tuned on large-scale multimodal instruction-following datasets~\cite{bai2023qwen, liu2024visual, schuhmann2022laion} obtained from the web. This significant breakthrough has garnered widespread attention from researchers in fields such as medical imaging~\cite{li2024llava} and human-computer interaction~\cite{chiang2024mobility}.

In the field of remote sensing, although deep learning methods~\cite{lobry2020rsvqa, kafle2016answer, zhang2023multi, bazi2022bi, chang2023changes, liu2023progressive, jin2022futh} have achieved significant success across various remote sensing multimodal tasks, previous approaches have largely been limited to task-specific models and even switching datasets within the same task can present challenges for these models~\cite{marino2019ok, lobry2020rsvqa, csahin2023deep, jin2022futh, liu2023decoupling}. This restricts the generalization performance of models when executing remote sensing multimodal tasks in real-world scenarios.
% , while also leading to inefficient use of computational resources.

% However, previous works~\cite{hu2023rsgpt, kuckreja2024geochat, zhan2024skyeyegpt} often design specialized model structures for specific tasks. It means new models need to be created when switching to a task involving another visual input type (or even a different dataset). This leads to resource wastage and limits the generalizability of VLMs, diverging from the original vision of developing AI assistants for remote sensing based on VLMs. The multimodal remote sensing tasks can be categorized into three types based on the form of visual input: single Image, dual-time image pair, and video, involving typical tasks such as visual question answering~\cite{lobry2020rsvqa}, change captioning~\cite{liu2022remote}, and video scene classification~\cite{mou2020era}, respectively. The VLM capabilities in the remote sensing field are still underexplored since previous VLMs are restricted to a few types of multimodal tasks and cannot unify different types of visual inputs in a single model. 

\begin{figure}[tb]
  \centering
   \includegraphics[width=1.0\linewidth]{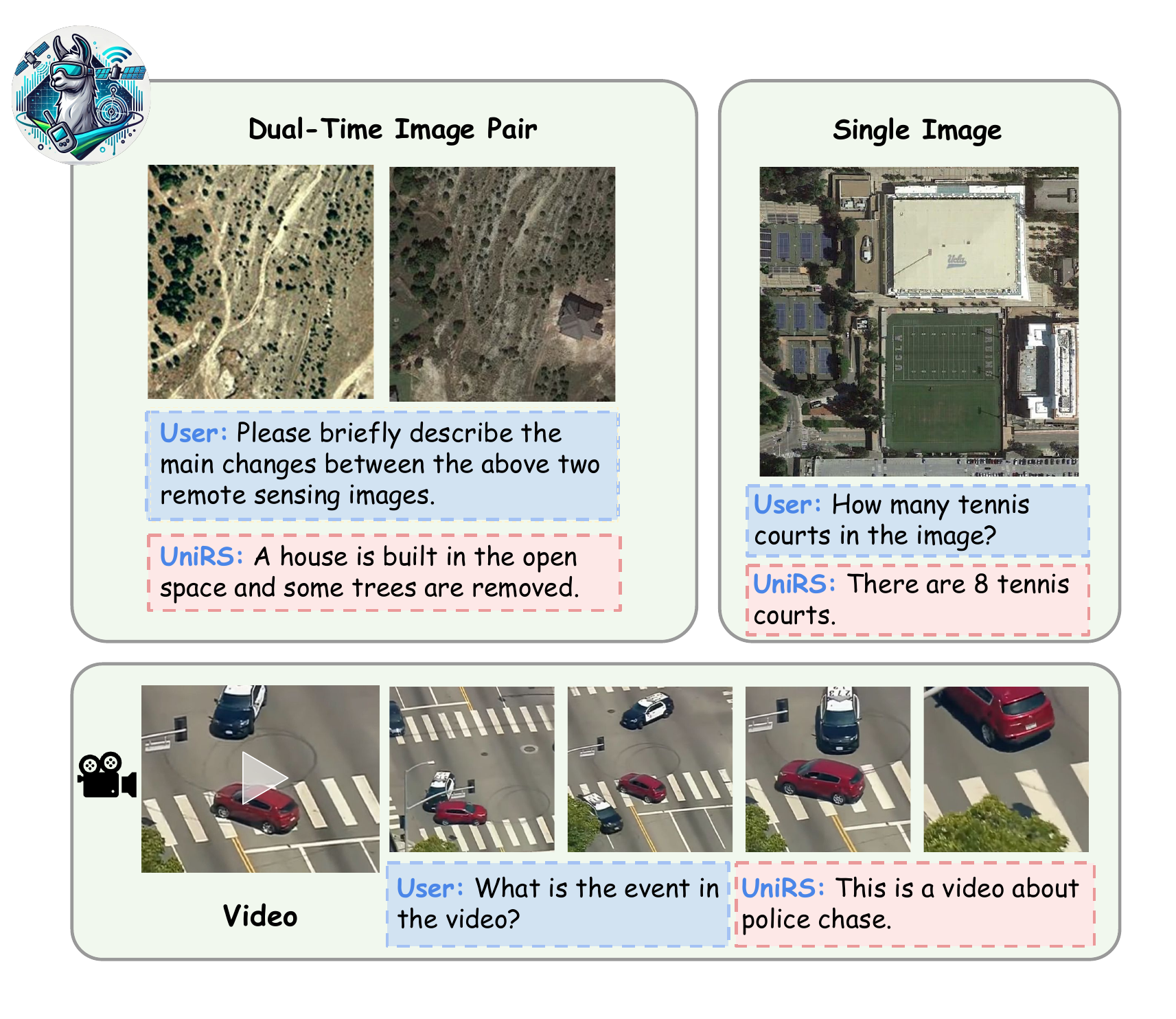}

   \caption{Our UniRS is a framework unifying multi-temporal remote sensing tasks of various visual inputs within a single model. It can analyze three critical types of remote sensing visual inputs i.e., single image, dual-time image pair, and video, under task instructions. Our research focuses on typical remote sensing tasks for each input type, including visual question answering, change captioning, and video classification.}
   \label{fig:teaser}
\vspace{-0.35cm}
\end{figure}

Recently, researchers in the remote sensing community have constructed large-scale remote sensing multimodal instruction-tuning datasets~\cite{muhtar2024lhrs, kuckreja2024geochat, zhan2024skyeyegpt} and established performance benchmarks to bridge the gap between general-purpose Vision-Language Models (VLMs) and remote sensing image analysis. These efforts have led to the development of remote sensing VLMs~\cite{ irvin2024teochat, bazi2024rs, muhtar2024lhrs, hu2023rsgpt}, such as GeoChat~\cite{kuckreja2024geochat} and EarthGPT~\cite{zhang2024earthgpt}, which demonstrate excellent generalization across various remote sensing multimodal tasks. However, these models offer limited exploration of the generalization boundaries of VLMs within remote sensing and lack the capability to process more diverse remote sensing visual inputs. GeoChat can follow instructions to execute multiple analytical tasks based on remote sensing images and has proposed large-scale multitask datasets and evaluation methods for multimodal task performance. EarthGPT, then, further addresses the diversity of remote sensing image inputs by incorporating optical, infrared, and SAR imaging modalities for analysis. However, EarthGPT still focuses primarily on single-image understanding. Data from single images, which lack inherent correlations, offer limited information for remote sensing knowledge extraction. In contrast, visual input with temporal relationships, such as dual-time image pairs and videos (or image sequences), can provide greater incremental value for remote sensing analysis and offer multimodal tasks with significant research potential.
% However, EarthGPT still focuses on the understanding of single images and 
% % does not fully consider the richness of input types in real-world open-domain remote sensing analysis scenarios. It 
% lacks the capability to handle dual-time image pairs and video (or image sequences), which are also critical types of remote sensing imagery. 
Therefore, there is an urgent need in the remote sensing field for a unified framework that integrates multi-temporal remote sensing tasks of multiple input types.

% To address the gap, we propose UniRS, the first vision-language model \textbf{Uni}fying multimodal \textbf{R}emote \textbf{S}ensing tasks of various visual input types. Different from previous VLMs for remote sensing with limited input type, our UniRS can handle visual question answering, change captioning, and video scene classification tasks simultaneously, which covers common visual inputs in remote sensing. We employ a unified visual representation for diverse multimodal remote sensing tasks, encoding various visual inputs~(e.g., single image, dual-time image pair, and video) into unified visual embeddings for the model to understand. To meet the high granularity requirements of the dual-time image pair understanding task which demands capturing local differences of interest and temporal relationships between images, we design a dedicated Change Extraction Module to capture the local spatiotemporal correlation features. This module enhances global spatial correlation features by incorporating dual-time cosine distance embeddings and fuses features from dual-time images to extract local spatiotemporal association features, enabling the model to discover the rich implicit semantics within dual-time image pairs. Furthermore, UniRS is jointly fine-tuned on a mixed dataset containing data of various types~(i.e., single image, dual-time image pair, video), leading the VLM to learn rich spatial and temporal features embedded in diverse visual data.

To bridge this gap, we propose UniRS, the first vision-language model \textbf{Uni}fying multi-temporal \textbf{R}emote \textbf{S}ensing tasks of various visual input types. Different from previous VLMs for remote sensing with only single image and text as input, our UniRS can handle single images, dual-time image pairs, and videos with textual inputs. This versatility enables UniRS to perform multiple tasks concurrently, including question answering, describing temporal changes, and classifying video scenes. Specifically, to enable the model to accept various types of visual inputs for multi-temporal remote sensing tasks, we adopt a unified visual embedding representation approach, allowing visual features from single images, dual-time image pairs, and videos to be aligned and merged with textual embeddings in a unified form. To meet the high granularity requirements of the dual-time image pair understanding tasks, we design a dedicated Change Extraction module for this type of visual input. This module incorporates cosine distance embeddings between dual-time features to enhance global spatial associations and further extracts local spatiotemporal correlation features through feature fusion, enhancing meaningful semantic changes occurring over time in image pairs within the context of global visual features.  Moreover, to leverage the excellent contextual understanding and instruction-following capabilities of the LLM foundation, we design a prompt augmentation mechanism. This mechanism uses the prior knowledge of the general-purpose VLM to initially interpret visual inputs and adds contextual clues to task instructions, optimizing UniRS's reasoning process. To promote knowledge sharing across tasks involving different temporal types, UniRS is jointly trained on a mixed dataset of various types, facilitating the learning of rich spatiotemporal features inherent in diverse visual inputs.

We develop our UniRS on three typical remote sensing multimodal tasks (i.e., visual question answering, change captioning, and video scene classification) of various temporal types and conduct extensive experiments. The results indicate that UniRS demonstrates exceptional generalization across various temporal visual inputs, surpassing SOTA expert models and other remote sensing VLMs. In the visual question answering task, under the RSVQA-HR~\cite{lobry2020rsvqa} test set, UniRS's zero-shot performance outperforms previous SOTA VLMs. Additionally, in the RSVQA-LR~\cite{lobry2020rsvqa} and CRSVQA~\cite{zhang2023multi}, UniRS also achieves SOTA under supervised settings. In the change captioning task, UniRS outperforms SOTA traditional methods on the LEVIR-CC~\cite{liu2022remote} and shows significant improvement over other VLMs. In the video scene classification task, UniRS significantly exceeds the performance of traditional classifier models on ERA~\cite{mou2020era} test set, showcasing the powerful understanding capabilities of VLMs for video inputs.

The contributions of our work are as follows:

\begin{itemize}
\item We propose UniRS, the first vision-language model designed to tackle multi-temporal remote sensing tasks, including visual question answering, change captioning, and video scene classification. It establishes a unified framework that combines three critical temporal visual input types in remote sensing i.e., single image, dual-time image pair, and video, broadening the capabilities of VLMs in remote sensing analysis, providing a paradigm for future research in multi-task integration within the remote sensing community.

\item We design a dedicated Change Extraction module, which enhances the comprehension of spatiotemporal semantic information in dual-time image pairs. This module incorporates a spatial feature enhancement component and a dual-time image feature fusion mechanism, enabling the model to detect and interpret local differences of interest and temporal relationships between two images. The module achieves high granularity in extracting and enhancing the spatiotemporal correlations of images, which is crucial for tasks requiring nuanced change detection.

\item We design a prompt augmentation mechanism for the inference process, which leverages the visual-language interaction capabilities of general VLM to enrich templated task instructions and provide task-specific clues for the UniRS in multimodal comprehension. During the clue generation, we design specific prompts for each type of remote sensing visual input. This mechanism utilizes the extensive prior knowledge of general-purpose VLM, facilitating the transfer of general knowledge to remote sensing analysis.

\item We develop a multi-task joint fine-tuning framework, designing task-specific prompt templates for different types of visual inputs to distinguish between tasks. UniRS is jointly trained on mixed datasets and the training promotes knowledge sharing across different tasks, enhancing the model's ability to understand the spatiotemporal features of remote sensing images compared to individual training. We extensively evaluate UniRS on visual question answering, change captioning, and video scene classification tasks, and it achieves state-of-the-art in all tasks, showcasing its versatility and effectiveness in tackling multi-temporal remote sensing challenges.
\end{itemize}

\section{Related Work}

\subsection{Vision-Language Models}
In recent years, with the development of large language models such as LLaMA~\cite{touvron2023llama} and GPT~\cite{brown2020language,achiam2023gpt}, Vision-Language Models (VLMs)~\cite{lin2024vila,luo2024zero,springstein2024visual} integrating visual features analysis with instruction-following capabilities have also gained widespread attention. VLMs are typically composed of three main components: \textbf{i)} A pre-trained large language model that is used to follow user instructions and respond based on provided visual features; \textbf{ii)} A pre-trained visual encoder, used for extracting features from visual inputs, often utilizing contrastive training foundation models like CLIP~\cite{radford2021learning} or SigLIP~\cite{zhai2023sigmoid}; \textbf{iii)} A transformation network between visual features and language embeddings, which translates visual features through linear layers or MLPs into a space that supports the understanding of language models. In this setup, token-level projectors merge the transformed visual features with command embeddings into LLMs (e.g., LLaVA~\cite{liu2024visual,liu2024improved}), while feature-level projectors require additional modules to facilitate deep interaction and integration between textual and visual functions (e.g., Flamingo~\cite{alayrac2022flamingo, li2023blip}). 

Aligning the visual feature space with the textual space, VLMs inherit the excellent multitask generalization ability and complex task reasoning capabilities of LLMs, which have introduced new research objectives in fields traditionally constrained by the ``one-task-one-model'' problem. Over time, VLMs have been widely adopted in research areas such as remote sensing~\cite{luo2024skysensegpt,bazi2024rs,zhang2024earthgpt} and biomedicine~\cite{li2024llava}, driving progress across various scientific disciplines.

\subsection{VLMs in Remote Sensing}
Deep learning methods have driven significant advancements in multimodal remote sensing tasks, such as visual question answering~\cite{lobry2020rsvqa}, image captioning~\cite{cheng2022nwpu} and change captioning~\cite{liu2022remote}. However, previous work required task-specific models, which typically exhibited limited generalization across different datasets. The success of VLMs has inspired researchers to propose large-scale remote sensing instruction-tuning datasets and remote sensing VLMs~\cite{luo2024skysensegpt,bazi2024rs,zhang2024earthgpt}. These models have demonstrated excellent performance in remote sensing multitask generalization and instruction following. RSGPT~\cite{hu2023rsgpt} is the first VLM in the remote sensing domain but requires separate fine-tuning for each task. GeoChat~\cite{kuckreja2024geochat} and EarthGPT~\cite{zhang2024earthgpt} emphasize the multitask handling capabilities, with EarthGPT additionally incorporating remote sensing analysis for optical, SAR, and infrared imaging modalities. However, these models focus solely on tasks for single-image inputs. SkyEyeGPT~\cite{zhan2024skyeyegpt} supports the video captioning task but does not explore more refined temporal analysis tasks such as multi-time point analysis tasks, which contain higher-density temporal change features than temporally coherent video analysis. TEOChat~\cite{irvin2024teochat}, on the other hand, emphasizes temporal image sequence understanding and maintains excellent generalization for single-image tasks, but does not support video-type inputs. 
% All of the above models, to varying extents, overlook the diversity of visual input forms in the remote sensing domain. 
Multi-temporal tasks like remote sensing visual question answering receive inputs from single images; video scene classification~\cite{jin2022futh,yang2022multiscale,shi2023alleviating} and video captioning tasks require video inputs; while tasks like change captioning~\cite{liu2023decoupling,chang2023changes,liu2022remote} often require dual-time image pair inputs for temporal sequence analysis. The previous models for remote sensing tasks do not properly support multi-temporal remote sensing visual information, although engineering can be conducted. To tackle this issue, we propose a model that unifies various multi-temporal tasks with multiple visual input types and conduct extensive instruction tuning, which works on various remote sensing tasks.

\subsection{Multi-task Learning} Multi-task learning~\cite{caruna1993multitask, caruana1997multitask} is a machine learning approach aimed at improving model generalization and training efficiency by leveraging shared representations across tasks. Multi-task learning primarily encompasses two paradigms: hard parameter sharing, where model parameters are shared across tasks, and soft parameter sharing, where each task uses its own model and knowledge transfer is achieved through regularization. In the fields of LLMs and VLMs, multi-task learning is also a key method for enhancing model performance~\cite{jaradat2024multitask, chen2023tigerbot, liu2024mftcoder}. It predominantly adopts the hard parameter sharing paradigm, where a single model is trained to perform multiple tasks, which enables the model, with its vast number of parameters and rich prior knowledge, to effectively learn shared features across related tasks, facilitating knowledge transfer between tasks and thereby improving the model's overall performance and robustness across various tasks.

Previous work about remote sensing VLM~\cite{kuckreja2024geochat, zhan2024skyeyegpt, zhang2024earthgpt} has predominantly adopted the hard parameter sharing paradigm, fine-tuning VLMs on multi-task datasets to achieve models with excellent generalization across various multimodal tasks. The goal of our work is to construct a 
multi-temporal unified framework integrating analytical capabilities of multiple remote sensing input types (i.e., single image, dual-time image pair, and video). We also adopt the hard parameter sharing paradigm and jointly train the VLM on a mixed dataset comprising various visual input types. This approach facilitates knowledge sharing across different tasks and enhances the generalization performance of UniRS across diverse visual inputs.

\section{Methodology}

\begin{figure*}[t]
  \centering
   \includegraphics[width=0.95\linewidth]{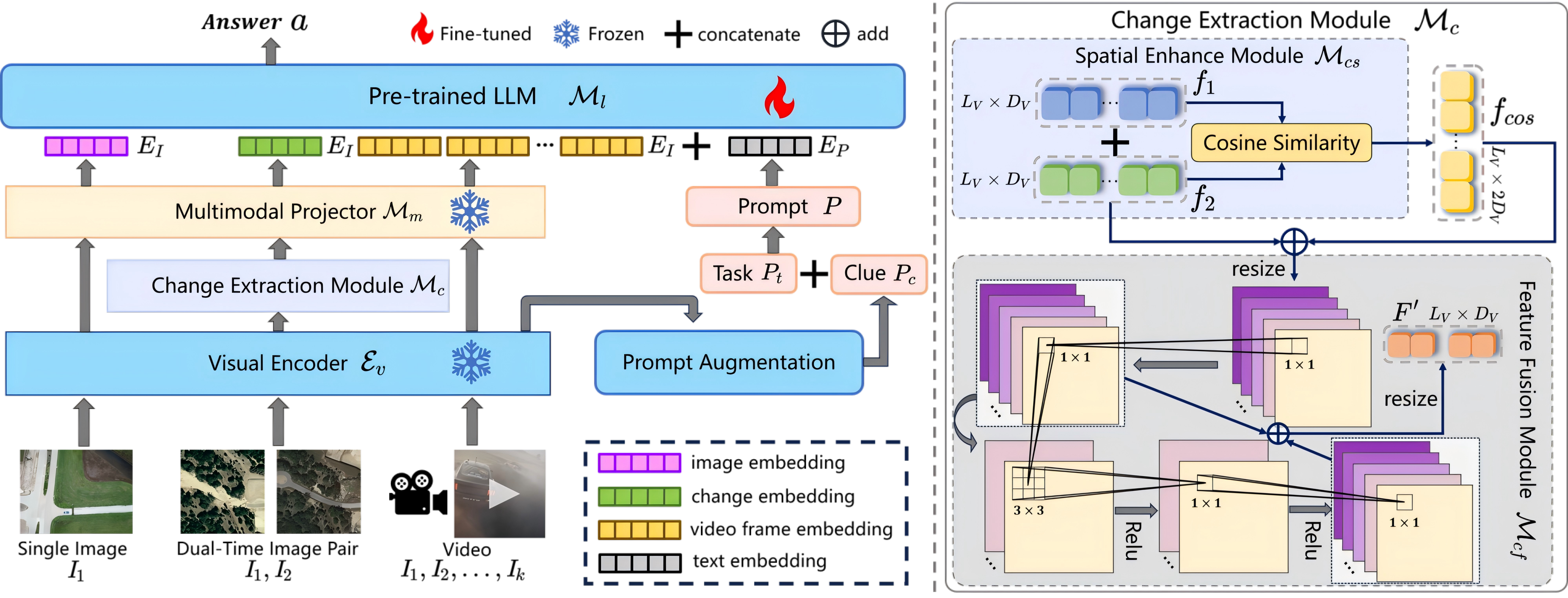}

   \caption{The architecture of our UniRS. The left part of this figure includes the prompt augmentation mechanism and UniRS main architecture. UniRS is primarily composed of four components, i.e., visual encoder $\mathcal{E}_v$, multimodal projector $\mathcal{M}_{m}$, language module $\mathcal{M}_{l}$, and change extraction module $\mathcal{M}_{c}$. Here change extraction module $\mathcal{M}_{c}$ is designed for the dual-time image pair input to extract and enhance spatiotemporal relationship features between image pairs. During inference, all visual inputs $\bm{I}$ are encoded into visual features $F$ by the visual encoder $\mathcal{E}_v$. In the prompt augmentation mechanism, initial visual clues $P_c$ are obtained after parsing and merged with the task instruction $P_t$ to form the full prompt $P$. In UniRS, the multimodal projector $\mathcal{M}_{m}$ projects visual feature $F$ into the text feature space as visual embedding $E_{I}$, which is then combined with the text embedding $E_{P}$ and fed into the language module $\mathcal{M}_{l}$ to get the final answer $a$. The right part of this figure is the structure of the change extraction module $\mathcal{M}_{c}$.
   }
   \label{fig:framework}
\vspace{-0.35cm}
\end{figure*}

\begin{figure*}[t]
  \centering
   \includegraphics[width=0.95\linewidth]{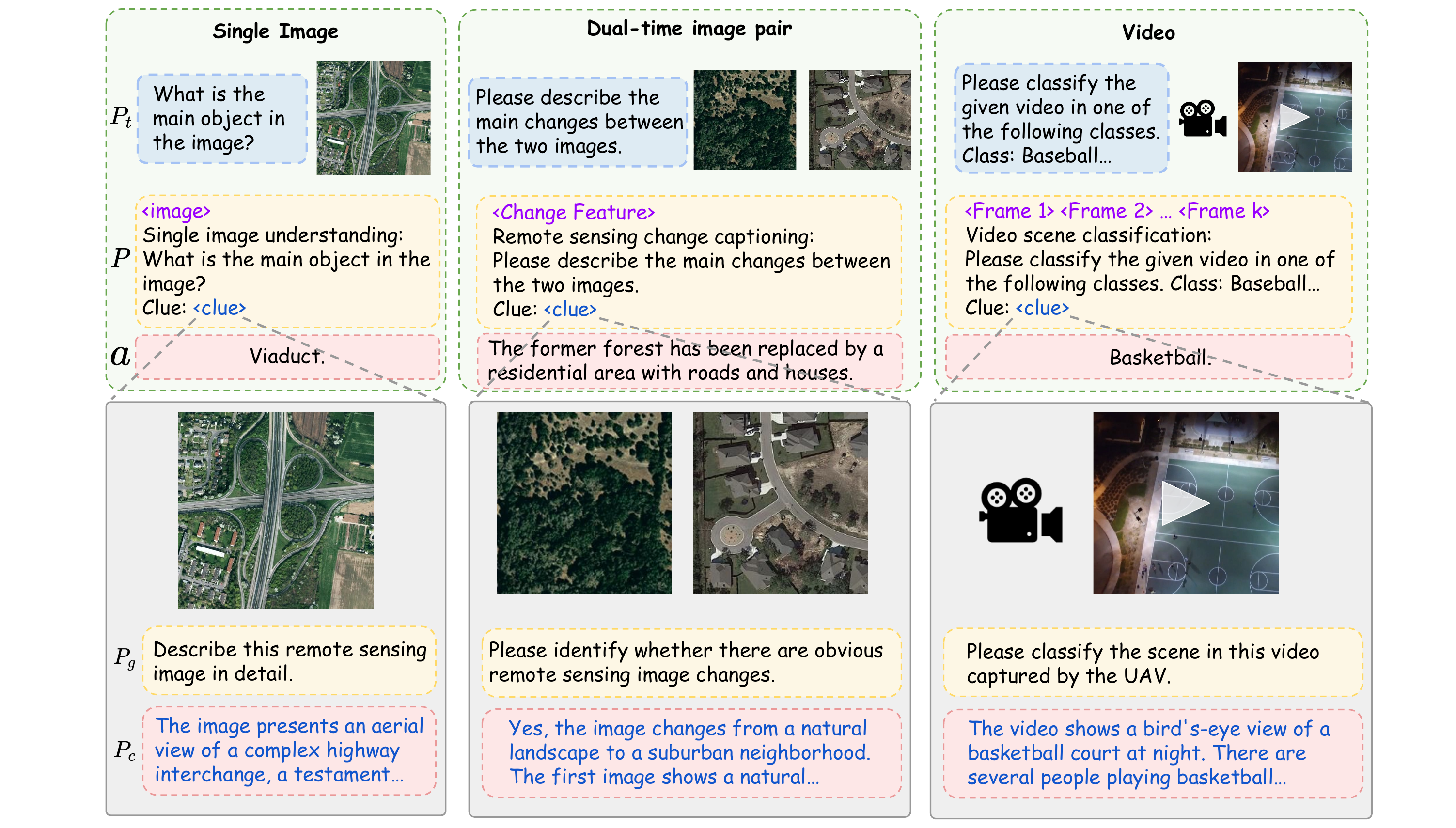}

   \caption{The inference process of UniRS using prompt augmentation mechanism. During the execution of remote sensing tasks, visual inputs are first processed by the base model, where clues $P_{c}$ are generated under the fixed prompts $P_g$ customized for each input type. These clues, special markers, task tags, and the task instruction $P_{t}$, are then merged to form the prompt $P$ input into UniRS. The model then generates the corresponding response $a$.
   }
   \label{fig:inference}
\vspace{-0.35cm}
\end{figure*}

We introduce the UniRS framework here. Section~\ref{method:A} provides an overview of UniRS, while section~\ref{method:B} details the individual modules within the framework. Section~\ref{Method:C} presents the joint training framework of our UniRS.

\subsection{Overview} \label{method:A}

In multi-temporal remote sensing tasks based on multiple visual inputs, users can provide the model with various forms of visual inputs $\bm{I}=\{I_1, \dots, I_k\}$ (i.e., single image, dual-time image pair, video), where $I_t$ is the $t$-th image or frame, along with corresponding instructions $P_{t}$ to receive appropriate answer $a$. Our UniRS integrates the ability to process three key 
multi-temporal remote sensing visual inputs (i.e., single image, dual-time image pair, and video) within a unified framework, and is studied on the following remote sensing tasks.
\begin{itemize}
    \item \textbf{Remote Sensing Visual Question Answering:} In this task, users provide a single image input $\bm{I}$ (i.e., $I_1$) and a corresponding instruction $P_{t}$. The model responds with answer $a$ appropriately after analyzing the image according to the instruction $P_{t}$.
    \item \textbf{Remote Sensing Change Captioning:} 
    % The user inputs are remote sensing image snapshots $I_1$ and $I_2$ of the same location but captured from different time.
    The input consists of two remote sensing images, $I_1$ and $I_2$, which capture the same geographical location at different timestamps.
    After encoding into feature $F$ by the visual encoder, the change extraction module processes the feature into $F'$, which is then projected to visual embedding $E_I$. LLM responds with answer $a$ according to $E_I$ and instruction $P_{t}$. The model discerns whether any changes of interest have occurred and describes the specific changes.
    \item \textbf{Remote Sensing Video Scene Classification:} This task's input is a video $\{I_1, \dots, I_k\}$ with $k$ frames. Based on the video input $\bm{I}$ and the corresponding instruction $P_{t}$, the model responds with a classification result $a$.
\end{itemize}

% UniRS is instruction fine-tuned based on the VILA 1.5 model using a mixed dataset of single image, dual-time image pair, and video. VILA1.5~\cite{lin2024vila} is a series of vision-language models trained on large-scale interleaved image-text data, with capabilities for general video understanding and Dual-Time image understanding. It follows a simple structural design similar to LLAVA, comprising three modules, i.e., visual encoder, multimodal projector, and large language model (LLM). Here the multimodal projector aims to project visual features into the text space. The key difference lies in UniRS's adaptation to the local similarity feature understanding required by dual-time image pair tasks. To address the unique needs of this visual input type, UniRS introduces a specialized Change Extraction module. The original dual-time image pair understanding capability of VILA 1.5 does not meet the needs of the remote sensing field for this visual input scenario. Additionally, we design a unified visual token representation method for all these visual inputs. 

\subsection{UniRS Framework} \label{method:B}
UniRS follows a similar simple architectural design to LLaVA~\cite{liu2024visual}, consisting of four modules: \textbf{i)} The visual encoder which extracts visual features; \textbf{ii)} The change extraction module which further enhances the spatiotemporal correlation features in dual-time image pairs; \textbf{iii)} The multimodal projector which projects visual features into the text space; \textbf{iv)} Large language model (LLM) as the foundation for interactive answering. To fully leverage the contextual reasoning capabilities of the LLM and assist UniRS in executing remote sensing tasks, we design a prompt augmentation mechanism. This mechanism utilizes the prior knowledge of the general-purpose VLM serving as the base model to provide UniRS with relevant clues. Additionally, UniRS is jointly fine-tuned on a mixed dataset to facilitate multi-task learning. An overview of our UniRS is shown in Fig.~\ref{fig:framework}.

\subsubsection{UniRS Architecture} \label{method:B.1}

In the process of remote sensing multimodal tasks, UniRS receives visual input $\bm{I}$ and prompt $P$ generated under prompt augmentation mechanism. Based on this input, UniRS generates the corresponding response $a$. 

UniRS uses VILA-1.5~\cite{lin2024vila}, a series of vision-language models trained on large-scale interleaved image-text data, as its base model, and a Change Extraction module has been specifically designed to handle dual-time image pair input tasks. The following section provides a detailed description of the model's architecture.

\textbf{Visual Encoder.} UniRS adopts the powerful visual language foundation model SigLIP~\cite{zhai2023sigmoid} as the visual encoder, formulated as $\mathcal{E}_v$. For a given visual input sequence $\bm{I} \in \mathbb{R}^{k \times 3 \times h \times w}$ (where $k$ equals 1 for the single image; $k$ equals 2 for dual-time images; $k$ is the number of video sampling frames for video inputs. $h$ and $w$ refer to the input size), the visual encoder encodes the visual input $\bm{I}$ into visual features $F \in \mathbb{R}^{kL_{V} \times D_V}$ ($L_{V}$ refers to the visual token length of 
a single image and $D_V$ is the dimension depth of the visual
feature):
\begin{equation}
	F = \mathcal{E}_v(\bm{I})\,.
\end{equation}
For input $\bm{I}=\{I_1, \dots, I_k\}$, $\mathcal{E}_v$ encode each image or frame $I_i$ into $f_i \in \mathbb{R}^{L_{V} \times D_V}$.

\textbf{Change Extraction Module.} Although the VILA-1.5 model can understand the temporal and spatial semantic relationships of visual tokens within context, the visual features $F$ generated solely by the visual encoder from visual input may not fully meet the demand for high-granularity analysis of local feature correlations between images in a dual-time image pair. Therefore, we have customized the change extraction module, formulated as $\mathcal{M}_{c}$, to cater to the specific needs of such visual input form, particularly for downstream tasks like change captioning.

As is shown in Fig.~\ref{fig:framework} (right), the module consists of the spatial enhance module, denoted as $\mathcal{M}_{cs}$, and the feature fusion module, denoted as $\mathcal{M}_{cf}$. For the visual features $\bm{F}=\{f_1, f_2\} \in \mathbb{R}^{2L_V \times D_V}$ derived from dual-time image pair input, $\mathcal{M}_c$ extracts and enhances meaningful spatiotemporal features while preserving the original semantic information. These features are encoded into a change feature map for further understanding. In the spatial enhance module $\mathcal{M}_{cs}$, we concatenate $f_1$ and $f_2$ to get $[f_1;f_2] \in \mathbb{R}^{L_V \times 2D_V}$, resize it to $\mathbb{R}^{2D_V \times \frac{h}{d_p} \times \frac{w}{d_p}}$ (the patch size of visual encoder is $d_p \times d_p$) and add cosine similarity distance embedding to enhance the local spatial correlation information between the dual-time image features as follows,
\begin{equation}
    [f_1';f_2'] = \mathcal{M}_{cs}(f_1, f_2)+[f_1;f_2]\,.
\end{equation}
$[\cdot;\cdot]$ denotes concatenation and $[f_1';f_2'] \in \mathbb{R}^{2D_V \times \frac{h}{d_p} \times \frac{w}{d_p}}$.

Subsequently, in the feature fusion module $\mathcal{M}_{cf}$, we design a three-layer 2D convolutional network with kernel sizes of $1\times1$, $3\times3$, and $1\times1$, respectively. ReLU is the activation function between layers, and residual connections are introduced. Our approach aims to fuse the visual features of the dual-time image pair, extracting local spatiotemporal features where changes of interest occur between the images and then enhance them on top of the global visual features background through the multi-layer 2D convolutional network:
\begin{equation}
    F' = Conv[f_1';f_2']+\mathcal{M}_{cf}(Conv[f_1';f_2'])\,,
\end{equation}
where $F' \in \mathbb{R}^{D_V \times \frac{h}{d_p} \times \frac{w}{d_p}}$ and $Conv$ denotes a single layer convolution with the kernel size of $1 \times 1$, halving the feature depth of output feature to $D_V$.
\begin{equation}
    \label{eq:ce}
    F' = \mathcal{M}_{c}(f_1, f_2)
\end{equation}
As shown in Equation~(\ref{eq:ce}) and Fig.~\ref{fig:framework} (right), the final visual feature map $F'$ is processed by $\mathcal{M}_{c}$, representing the spatiotemporal correlation enhanced features.

\textbf{Multimodal Projector.} The visual encoder and Change Extraction module encode the visual input into feature $F \in \mathbb{R}^{kL_V \times D_V}$, while the multimodal projector, formulated as $\mathcal{M}_m$, maps the feature from the visual feature space to the word embedding space suitable for LLM input:
\begin{equation}
    E_I = \mathcal{M}_m(F)\,,
\end{equation}
where $E_I \in \mathbb{R}^{kL_d \times D_P}$ ($L_{d}$ is the feature length after downsampling and $D_P$ is the depth of textual embedding).

The Multimodal Projector consists of a downsampling module, $\mathcal{M}_{md}$, and a MLP network. Visual features are first downsampled and then passed into the MLP network for dimensional transformation. In the $\mathcal{M}_{md}$, the visual features are downsampled to $F_d \in \mathbb{R}^{k L_{d} \times 4D_V}$($L_{d}=\frac{L_V}{4}$),
\begin{equation}
    F_d = \mathcal{M}_{md}(F)\,,
\end{equation}
which helps aggregate local features and extract high-level semantic features while significantly reducing the length of visual tokens. Then, the MLP network maps the visual features $F_d$ to the word embedding space, converting them into visual embeddings $E_I \in \mathbb{R}^{kL_{d} \times D_P}$,
\begin{equation}
    E_I = MLP(F_d)\,,
\end{equation}
where $D_P$ is the dimension depth of LLM textual embedding.

\textbf{Large Language Model.} UniRS utilizes the open-source Sheared-LLaMA (3B)~\cite{xia2023sheared} as the language decoder. This is a lightweight LLM derived from LLaMA 2 through structured pruning, resulting in a reduced number of parameters, allowing us to quickly and cost-effectively fine-tune it on remote sensing multimodal tasks. During the multimodal task inference, the input prompt $P$ generated by the prompt augmentation mechanism is encoded into text embeddings $E_P \in \mathbb{R}^{rL_P \times D_P}$ ($r$ refers to the input sequence length) by the tokenizer, $\mathcal{M}_{p}$,
% \begin{equation} \label{equation:Prompt}
%     P = Template(P_{c}, P_{t})\,,
% \end{equation}
\begin{equation}
    E_P = \mathcal{M}_p(P)\,.
\end{equation}
The textual embeddings $E_P$ are then concatenated with the visual embeddings $E_I$ output by the multimodal projector and input into the LLM, denoted as $\mathcal{M}_{l}$. The LLM generates a response $a$ corresponding to the prompt and visual input,
\begin{equation}
    a = \mathcal{M}_{l}([E_I;E_P])  \,.
\end{equation}

\subsubsection{Prompt Augmentation Mechanism} 
% \wenjia{I would recommend to first introduce UniRS Architecture, then introduce the prompt augmentation mechanism.}
The prompt augmentation mechanism utilizes the frozen base model of UniRS (i.e., VILA-1.5) to provide an initial understanding of the input remote sensing images, which serves as clues for UniRS when performing multi-temporal remote sensing tasks. This mechanism is designed based on the fundamental understanding capabilities of general vision-language models for remote sensing images. 

Here, the task instruction provided by the user during the execution of multi-temporal remote sensing tasks is $P_t$, and the clue generated by the base model is $P_c$, where the prompt guiding the base model to generate the clue is denoted as $P_g$. The final prompt input to UniRS is denoted as $P$, which is obtained by merging the $P_t$ with $P_c$ in a template.

During inference of UniRS, the frozen base model, detailed in Section~\ref{method:B.1} and denoted as $\mathcal{M}_{b}$, receives visual input $I$ and the task-based prompt $P_{g}$ to generate clues in a zero-shot setting. The model then outputs the corresponding clues $P_{c}$:
\begin{equation}
    P_{c} = \mathcal{M}_{b}(I,P_{g})\,.
\end{equation}
The visual clues $P_{c}$ output by $\mathcal{M}_{b}$ are concatenated with the text instructions $P_{t}$ for the remote sensing multimodal tasks using a templated format, resulting in the combined prompt $P$,
\begin{equation} \label{equation:Prompt}
    P = Template(P_{c}, P_{t})\,,
\end{equation}
which is then input into UniRS. The responses $P_{c}$ generated by $\mathcal{M}_{b}$ contain clues beneficial for multi-temporal remote sensing tasks. By adding these clues $P_c$ to the task instructions $P_{t}$, the mechanism not only provides contextual clues to support the inference of UniRS but also enriches the instructions, thus augmenting the prompt for the model.

\subsection{Joint Instruction Tuning} \label{Method:C}
This section provides a detailed illustration of the joint training framework and the design introduced to support joint training with multi-temporal data.

% \begin{figure}[t]
%   \centering
%   % \fbox{\rule{0pt}{2in} \rule{0.9\linewidth}{0pt}}
%    \includegraphics[width=0.9\linewidth]{Instruction_Template.pdf}

%    \caption{Task-based visual language instruction template for joint tuning}
%    \label{fig:Instruction}
% \vspace{-0.35cm}
% \end{figure}

\begin{table}[tb]
\centering
\caption{The instructions $P_t$ of UniRS for different training datasets.
}
\label{tab:instruction_template}
\renewcommand{\arraystretch}{1.2} \resizebox{0.48\textwidth}{!}{

\begin{tabular}{c|l}
\hline \rowcolor{gray!15}
\textbf{Dataset} & \multicolumn{1}{c}{\textbf{Task Instruction} $P_t$}                                             \\ \hline
GeoChat-Instruct~\cite{kuckreja2024geochat} &
  \begin{tabular}[c]{@{}l@{}}Is there a water area on the right of a road? Answer in one word or a short phrase.\\ Classify the given image in one of the following classes. Classes: airplane, parking lot...\\ ... (directly use the instructions in the dataset)\end{tabular} \\ \hline
LEVIR-CC~\cite{liu2022remote}         & Please identify whether there are obvious remote sensing image changes.                   \\ \hline
ERA~\cite{mou2020era}              & Classify the given video in one of the following classes. Classes:Baseball, Basketball... \\ \hline
\end{tabular}}
\vspace{-0.35cm}
\end{table}

\begin{table}[tb]
\centering
\caption{Prompt $P_g$ used with diverse visual input in clue generation.
}
\label{tab:clue_generation_template}
\renewcommand{\arraystretch}{1.2} \resizebox{0.48\textwidth}{!}{

\begin{tabular}{c|l}
\hline \rowcolor{gray!15}
\textbf{Visual Input Type}    & \multicolumn{1}{c}{\textbf{Prompt} $P_g$}                              \\ \hline
Single image & Describe this remote sensing image in detail.                \\ \hline
\begin{tabular}[c]{@{}c@{}}Dual-time\\ image pair\end{tabular} & Please identify whether there are obvious remote sensing image changes. \\ \hline
Video        & Please classify the scene in this video captured by the UAV. \\ \hline
\end{tabular}}
\vspace{-0.35cm}
\end{table}

\textbf{Task-based Unified Visual Representation.}\label{section:viusal_representation}
Our UniRS employs a task-based unified visual token representation scheme that can handle multi-temporal tasks of the three critical visual input forms in remote sensing i.e., single image, dual-time image pair, and video. When generating visual clues, we follow the multi-image processing method natively supported by VILA-1.5, concatenating the visual features encoded by the visual encoder based on the image or video frame number. In UniRS, we maintain the representation methods for single images and videos, but we have also customized a specific representation method for the image-pair input mode to suit the special design requirements of UniRS when handling dual-time image pairs.
% To enable the model to flexibly receive and understand three critical types of visual input (i.e. single image, dual-time image pair, and video) under various tasks in remote sensing and engage in vision-language dialogue, we adopt a unified representation scheme that adapts these visual inputs to the LLM input embedding format.

For visual question answering (single image), we mark the visual input with a special marker $\langle$\textit{image}$\rangle$ in the textual instructions and replace it with the single image visual embedding $E_{I}^{single} \in R^{L_{d} \times D_P}$ in the LLM’s input embedding. For change captioning (dual-time image pair), we still use a single special marker $\langle$\textit{Change Feature}$\rangle$ in the instruction, replacing it with the spatiotemporal correlation feature embedding $E_{I}^{dual} \in R^{L_{d} \times D_P}$. For video scene classification (video), depending on the number of sampled frames $k$, we add $k$ consecutive markers $\langle$\textit{Frame K}$\rangle$ in the text instructions and replace them sequentially with the visual embeddings $E_{I}^{video} \in R^{NL_{d} \times D_P}$ according to the sampling sequence. During inference, the visual embeddings are concatenated with the text embeddings in a unified format and fed into the LLM.

\textbf{Data Organization.} The dataset used for instruction-tuning of UniRS is a mixture of three commonly used remote sensing datasets: GeoChat-Instruct~\cite{kuckreja2024geochat}, LEVIR-CC~\cite{liu2022remote}, and ERA Dataset~\cite{mou2020era}. GeoChat-Instruct integrates three object detection datasets (i.e., DOTA~\cite{xia2018dota}, DIOR~\cite{cheng2022anchor}, FAIR1M~\cite{sun2022fair1m}), one scene classification dataset (i.e., NWPU-RESISC-45~\cite{cheng2017remote}), and two visual question answering datasets (i.e., LRBEN~\cite{lobry2020rsvqa}, Floodnet~\cite{rahnemoonfar2021floodnet}). It is a high-quality instruction-tuning dataset generated using Vicuna-adapted text context prompts, containing 306k image-instruction pairs for training. LEVIR-CC is a remote sensing change captioning dataset based on LEVIR-CD~\cite{chen2020spatial}, manually annotated to include 10,077 dual-time image pairs, split into training (67.6\%), validation (13.2\%) and test (19.2\%) sets. Each pair of images is associated with five captions and the training set is used for joint training. The ERA is a UAV overhead-view video classification dataset with 2,864 videos crawled from YouTube, classified into 25 labels. The dataset is split into training (51.4\%) and test (48.6\%) sets, with the training set used for instruction fine-tuning.

\textbf{Prompt Construction.} The final prompt $P$ input to UniRS consists of four components: \textbf{i)} visual input marker $\langle$\textit{img}$\rangle$ used as a placeholder for visual features, \textbf{ii)} task tag to distinguish between different remote sensing tasks, \textbf{iii)} the task instruction $P_t$ provided by the user, and \textbf{iv)} the clue $P_c$ generated by the prompt augmentation mechanism. Examples of the prompt are illustrated in Fig.~\ref{fig:inference}. The visual token representation method has been introduced earlier in Section~\ref{section:viusal_representation}. To distinguish between different types of tasks, we add task tags, i.e. ``Single image understanding:'', ``Remote sensing change captioning:'' and ``Video scene classification:''. In $P_t$, since the LEVIR-CC and ERA datasets were originally developed for expert model training and did not natively support VLM instruction-tuning, we design task-specific instructions for them, as shown in TABLE~\ref{tab:instruction_template}. Additionally, within the prompt augmentation mechanism, we design fixed prompts $P_g$ for each input type to generate clues, as shown in TABLE~\ref{tab:clue_generation_template}. Finally, the clue is incorporated into the prompt in the format ``Clue: $\langle$\textit{clue}$\rangle$''.

\textbf{Joint Tuning Strategy.} 
% By adopting a unified representation scheme for different forms of visual input, we can conveniently perform instruction fine-tuning on UniRS using the mixed dataset across all three visual types. 
% Our UniRS is developed based on the VILA-1.5~\cite{lin2024vila} model, with pre-trained model parameters imported for all modules except the Change Extraction module. 
We mix the datasets of all three visual types to train our UniRS, which helps to obtain a unified representation for multi-temporal visual input. The UniRS adopts VILA-1.5~\cite{lin2024vila} as the base model, and we design the change extraction module to strengthen its abilities for different tasks. 
During joint training, the visual encoder of UniRS, the Change Extraction module, and the multimodal projector are all frozen, with only the LLM being fully fine-tuned using cross-entropy loss and the training objective is:
\begin{equation}
    \mathcal{L}(\theta)=-\sum_{t=1}^{L_{seq}} \log P\left(\mathbf{y}_{i} \mid \mathbf{y}_{<i}, \mathbf{x} ; \theta\right)\,,
\end{equation}
where $\theta$ is the parameters of UniRS, $L_{seq}$ represents the length of the multimodal sequence, $\mathbf{x}$ is the input sequence and $\mathbf{y}_{i}$ is the $i$-th output word.

% By employing a unified representation scheme for different forms of visual input, we can conveniently instruction-tune UniRS on mixed datasets containing all three types of visual inputs. It is important to note that UniRS is developed based on the VILA 1.5 (3B) model. Except for the Change Extraction Module, all other modules are initialized with pre-trained model parameters obtained from large-scale datasets. We then instruction-tune the model on remote sensing vision-language datasets using these pre-trained parameters. During the joint training process, the Visual Encoder, Change Extraction Module, and Multimodal Projector are frozen, and only the LLM is fully fine-tuned using cross-entropy loss.

\section{Experiments}

% In this section, we conduct extensive experiments on three important types of visual input in the field of remote sensing (i.e., single image input, dual-time image pair, and video) under their respective typical vision-language tasks, specifically visual question answering, change captioning, and video classification.

We conduct experiments to validate the generalization performance of UniRS in multi-temporal remote sensing tasks across various input types. We first provide the specific implementation details of UniRS. Then we evaluate the UniRS across three main tasks, specifically visual question answering, change captioning, and video scene classification.
% , which represent the three types of remote sensing visual input (i.e., single image, dual-time image pair, and video).

\subsection{Implementation Details}\label{implementation}

% We first initialize the Visual Encoder, Multimodal Projector, and LLM using the parameters from VILA1.5-3B. To enable the Change Extraction Module to extract spatiotemporal correlation information embedded in dual-time image pairs effectively, we follow the work of Chg2Cap~\cite{chang2023changes}. Specifically, we use Siglip, employed in UniRS, as the encoder, with the Change Extraction Module extracting spatiotemporal correlations to generate visual feature embeddings which are then fed into a single-layer multi-head Transformer network, serving as the decoder, to generate captions. We pre-train the Change Extraction Module with this architecture, using the AdamW optimizer along with a StepLR learning rate scheduler, where the step size was set to 5 and gamma to 0.5. The maximum learning rate is 1e-4, with a batch size of 64. The pre-trained parameters are used to initialize the Change Extraction Module for joint instruction-tuning. During this stage, all modules except the LLM were frozen, and the LLM is fully fine-tuned. We use the AdamW optimizer with a cosine learning rate scheduler, setting the warmup ratio to 0.03 and the maximum learning rate to 1e-4. The fine-tuning is conducted for one epoch on the mixed dataset, with a batch size of 128. Both training stages are executed on four NVIDIA RTX4090 GPUs.

Our UniRS is initialized using VILA-1.5 (3B), specifically adopting the pre-trained SigLIP-so400m~\cite{zhai2023sigmoid}, MLP, and Sheared-LLAMA (3B)~\cite{xia2023sheared} models from the VILA-1.5 (3B) framework~\cite{lin2024vila}. To enable the Change Extraction module to capture the spatiotemporal features inherent in dual-time image pairs, we follow the work of Chg2Cap~\cite{chang2023changes}. Specifically, SigLIP serves as the encoder, while the Change Extraction module generates spatiotemporal feature embeddings
then fed into a single-layer multi-head Transformer network to generate captions. The Change Extraction module is pre-trained within this architecture and then used to initialize the module in UniRS. During subsequent joint instruction fine-tuning, parameters of the other modules are frozen, and only the LLM module undergo full fine-tuning on the mixed datasets (i.e., GeoChat-Instruct~\cite{kuckreja2024geochat}, LEVIR-CC~\cite{liu2022remote}, and ERA~\cite{mou2020era}).

In the first training process, we freeze the visual encoder and follow the hyperparameter settings of Chg2Cap~\cite{chang2023changes} for training the remaining modules. In the second process of joint instruction-tuning, we set the maximum training sequence length to 4096 and use the AdamW optimizer with a cosine learning rate scheduler, setting the maximum learning rate to 1$\times$10$^{-4}$, minimum learning rate to 0, and a warm-up ratio of 0.03. The model is trained for one epoch on the mixed dataset with a batch size of 128, in a total of 3,858 steps. Both training stages are executed on four NVIDIA RTX4090 GPUs.

\subsection{Remote Sensing Visual Question Answering}

\begin{table}[tb]
\centering
\caption{Comparison of the visual question answering performance on RSVQA-LR Dataset~\cite{lobry2020rsvqa}. VILA-1.5~\cite{lin2024vila} is evaluated under the zero-shot setting. Our UniRS, SkyEyeGPT~\cite{zhan2024skyeyegpt}, LHRS-Bot~\cite{muhtar2024lhrs} and Geochat~\cite{kuckreja2024geochat} are non-expert models. UniRS (further training) is compared with expert models.}
\label{tab:rsvqa-lrben}
\renewcommand{\arraystretch}{1.2} \resizebox{0.5\textwidth}{!}{
\begin{tabular}{cc|cccc}
\hline
\multicolumn{2}{c|}{\textbf{Method}}             & \textbf{Presence} & \textbf{Comparison} & \textbf{Rural/Urban} & \textbf{Avg. Accuracy} \\ \hline
\multicolumn{1}{c|}{Base}                        & VILA-1.5 (3B)~\cite{lin2024vila}  & 68.49       & 64.99       & 64.00          & 66.44       \\ \hline
\multicolumn{1}{c|}{\multirow{4}{*}{Expert}}     & RSVQA~\cite{lobry2020rsvqa}          & 87.47       & 81.5        & 90.00          & 86.32       \\
\multicolumn{1}{c|}{}                            & Bi-Modal~\cite{bazi2022bi}       & 91.06       & 91.16       & 92.66       & 91.63       \\
\multicolumn{1}{c|}{}                            & SHRNet~\cite{zhang2023spatial}         & 91.03       & 90.48       & 94.00          & 91.84       \\
\multicolumn{1}{c|}{}                            & RSGPT~\cite{hu2023rsgpt}          & 91.17       & 91.7        & \textbf{94.00} & \underline{92.29} \\ \hline
\multicolumn{1}{c|}{\multirow{3}{*}{Non-expert}} & SkyEyeGPT (7B)~\cite{zhan2024skyeyegpt} & 88.93       & 88.63       & 75.00          & 84.19       \\
\multicolumn{1}{c|}{}                            & LHRS-Bot (7B)~\cite{muhtar2024lhrs}  & 89.07       & 88.51       & 90.00          & 89.19       \\
\multicolumn{1}{c|}{}                            & GeoChat (7B)~\cite{kuckreja2024geochat}   & 91.09       & 90.33       & 94.00          & 90.70        \\ \hline \rowcolor{blue!15}
\multicolumn{1}{c|}{\multirow{1}{*}{Ours}}       & UniRS          & \underline{91.64} & \underline{92.68} & 90.00          & 92.21       \\
\rowcolor{blue!15}
\multicolumn{1}{c|}{\multirow{1}{*}{Ours}} & UniRS (further training) & \textbf{91.81}    & \textbf{93.23}      & \underline{93.00}             & \textbf{92.63}         \\ \hline
\end{tabular}}
% \vspace{-0.35cm}
\end{table}

We begin by introducing the datasets and experimental settings. Next, we provide a numerical analysis of UniRS across different datasets. Finally, we conduct a qualitative analysis to highlight the performance characteristics of UniRS.

% \textbf{Dataset.} We use the test sets of the RSVQA-LR~\cite{lobry2020rsvqa}, RSVQA-HR~\cite{lobry2020rsvqa} and CRSVQA~\cite{zhang2023multi} datasets for the quantitative evaluation of the visual question answering task. The LRBEN test set contains 10,004 question-answer pairs, covering four kinds of questions (i.e., presence, comparison, rural/urban classification, and counting). HRBEN contains two test sets, test-1 and test-2, and we selected test-2, which includes 62,554 question-answer pairs with three types of questions (i.e., presence, comparison and counting). During the evaluation, for the sake of fairness in comparison, as with GeoChat~\cite{kuckreja2024geochat}, we also exclude counting and area-related questions.

\textbf{Dataset.} We use the test sets from the RSVQA-LR~\cite{lobry2020rsvqa}, RSVQA-HR~\cite{lobry2020rsvqa}, and CRSVQA~\cite{zhang2023multi} datasets for quantitative testing of the RSVQA task. The RSVQA-LR dataset contains 772 remote sensing images of size 256$\times$256 and 77,232 question-answer pairs, of which 11.1\% of the data is split for testing. The RSVQA-HR dataset includes 10,659 images of size 512$\times$512 and 1,066,316 question-answer pairs, divided into test set 1 (20.5\%) and test set 2 (6.8\%). The CRSVQA dataset contains 4,639 remote sensing images of size $600\times600$ and 4,644 manually annotated question-answer pairs, with a test set consisting of 1,000 data points. For the testing of RSVQA-LR and RSVQA-HR, we follow the GeoChat benchmark~\cite{kuckreja2024geochat} settings, using test set 2 from RSVQA-HR and excluding counting and area-related questions. Additionally, we follow the setting of MQVQA~\cite{zhang2023multi}, adopting 10\% of the data as the test set and evaluating the performance under supervised assessment.

\begin{table}[tb]
\centering
\caption{Comparison of the zero-shot visual question answering performance on RSVQA-HR~\cite{lobry2020rsvqa} dataset. We compare our UniRS with general VLMs and remote sensing VLMs under zero-shot settings.}
\label{tab:rsvqa-hrben}
\renewcommand{\arraystretch}{1.2} \resizebox{0.5\textwidth}{!}{
\fontsize{5pt}{6pt}\selectfont
\begin{tabular}{cc|ccc}
\hline
\multicolumn{2}{c|}{\textbf{Method}}                & \textbf{Presence}       & \textbf{Comparison}     & \textbf{Avg. Accuracy}  \\
\hline
\multicolumn{1}{c|}{\multirow{3}{*}{General VLM}}   & VILA-1.5 (3B)~\cite{lin2024vila}            & 61.44          & 63.06          & 62.79          \\
\multicolumn{1}{c|}{}   &   MiniGPTv2~\cite{chen2023minigpt}    & 40.79     & 50.91     & 46.46     \\
\multicolumn{1}{c|}{}   &   LLaVA-1.5~\cite{li2024llava}        & \textbf{68.23}     & 65.45     & 66.67     \\ \hline
\multicolumn{1}{c|}{\multirow{2}{*}{RS VLM}}    &   GeoChat~\cite{kuckreja2024geochat}               & 59.02          & \underline{83.16}          & \underline{72.53}          \\
\multicolumn{1}{c|}{}   &   EarthGPT~\cite{zhang2024earthgpt}       & \underline{62.77}     & 79.53     & 72.06     \\  \hline
\rowcolor{blue!15}
\multicolumn{1}{c|}{\multirow{1}{*}{Ours}}   &   UniRS & 59.29 & \textbf{84.05} & \textbf{73.15} \\
\hline
\end{tabular}}
\vspace{-0.35cm}
\end{table}

\begin{table}[t]
\centering
\caption{Comparison of the visual question answering performance on CRSVQA~\cite{zhang2023multi} dataset. VILA-1.5~\cite{lin2024vila}, EarthGPT~\cite{zhang2024earthgpt}, GeoChat~\cite{kuckreja2024geochat}, and UniRS are tested with supervised settings.
}
\label{tab:crsvqa}
\renewcommand{\arraystretch}{1.1} \resizebox{0.35\textwidth}{!}{
\fontsize{4pt}{5pt}\selectfont
\begin{tabular}{cc|c}
\hline
\multicolumn{2}{c|}{\textbf{Method}}      & \textbf{OA}             \\
\hline
\multicolumn{1}{c|}{\multirow{1}{*}{Base}}   &VILA-1.5 (3B)~\cite{lin2024vila}       & 80.33     \\ \hline
\multicolumn{1}{c|}{\multirow{5}{*}{Expert}}   &Qonly~\cite{marino2019ok}                & 23.49          \\
\multicolumn{1}{c|}{}   &RSVQA~\cite{lobry2020rsvqa}                & 58.96          \\
\multicolumn{1}{c|}{}   &RSVQA(GRU)~\cite{lobry2020rsvqa}           & 59.41          \\
\multicolumn{1}{c|}{}   &SAN~\cite{kafle2016answer}                  & 61.17          \\
\multicolumn{1}{c|}{}   &MQVQA~\cite{zhang2023multi}                & 70.91          \\ \hline
\multicolumn{1}{c|}{\multirow{2}{*}{RS VLM}}   &EarthGPT~\cite{zhang2024earthgpt}             & 82.00        \\
\multicolumn{1}{c|}{}   &GeoChat~\cite{kuckreja2024geochat}      &   \underline{82.50}       \\
\hline \rowcolor{blue!15}
\multicolumn{1}{c|}{\multirow{1}{*}{Ours}}   &UniRS & \textbf{86.67} \\
\hline
\end{tabular}}
\vspace{-0.35cm}
\end{table}

\begin{table}[tb]
\centering
\caption{Comparison of the change captioning performance on LEVIR-CC Dataset~\cite{liu2022remote}. LLaVA-1.5~\cite{liu2024visual} and GeoChat~\cite{kuckreja2024geochat} refer to the engineered models fine-tuned on LEVIR-CC~\cite{liu2022remote} train set.}
\label{tab:levir_cc}
\renewcommand{\arraystretch}{1.2} \resizebox{0.35\textwidth}{!}{
\fontsize{5pt}{6pt}\selectfont
\begin{tabular}{cc|c}
\hline
\multicolumn{2}{c|}{\textbf{Method}}       & \textbf{CIDEr-D}         \\
\hline
\multicolumn{1}{c|}{\multirow{1}{*}{Base}} &VILA-1.5 (3B)~\cite{lin2024vila}   & 6.22             \\   \hline
\multicolumn{1}{c|}{\multirow{4}{*}{Expert}} &RSICCFormer~\cite{liu2022remote}  & 131.40          \\
\multicolumn{1}{c|}{} &PSNet~\cite{liu2023progressive}         & 132.62 \\
\multicolumn{1}{c|}{} &PromptCC~\cite{liu2023decoupling}     & 136.44          \\
\multicolumn{1}{c|}{} &Chg2Cap~\cite{chang2023changes}      & \underline{136.61}          \\
\hline
\multicolumn{1}{c|}{\multirow{2}{*}{Fine-tuned VLM}} &LLaVA-1.5~\cite{liu2024visual}      & 126.25    \\
\multicolumn{1}{c|}{} &GeoChat~\cite{kuckreja2024geochat}      & 128.36    \\ \hline
\rowcolor{blue!15}
\multicolumn{1}{c|}{\multirow{1}{*}{Ours}} &UniRS & \textbf{139.12} \\
\hline
\end{tabular}}
\vspace{-0.35cm}
\end{table}

\textbf{Quantitative Results.} \subsubsection{Comparision on RSVQA-LR} We compare UniRS with three groups of models on the RSVQA-LR test set: expert models that underwent multiple epochs of supervised training on RSVQA-LR instruction-answer pairs, i.e., RSVQA~\cite{lobry2020rsvqa}, Bi-Modal~\cite{bazi2022bi}, SHRNet~\cite{zhang2023spatial}, and RSGPT~\cite{hu2023rsgpt}; non-expert models fine-tuned on a mixed dataset, i.e., GeoChat~\cite{kuckreja2024geochat}, LHRS-Bot~\cite{muhtar2024lhrs}, and SkyEyeGPT~\cite{zhan2024skyeyegpt}; and general VLM base model of UniRS, i.e., VILA-1.5 (3B)~\cite{lin2024vila}. Additionally, since RSGPT~\cite{hu2023rsgpt} is trained for 5 epochs on the RSVQA-LR~\cite{lobry2020rsvqa} training set for testing, we also include a further training experiment group. Given that UniRS's training dataset includes LR, we train it for an additional 4 epochs. 

As shown in TABLE~\ref{tab:rsvqa-lrben}, compared to the non-expert model group, UniRS shows a significant improvement over previous work, achieving an average accuracy of 92.21\%, outperforming GeoChat~\cite{kuckreja2024geochat} by 1.51\%, with notable improvements in both ``presence'' and ``comparison'' categories, reaching 91.64\% and 92.68\%, respectively. It has only 3B parameters, less than half of GeoChat's 7B parameters. When compared to the expert model group, the further trained UniRS achieves state-of-the-art performance, with an average accuracy of 92.63\%, outperforming the previous SOTA, RSGPT~\cite{hu2023rsgpt}. After further training, UniRS reaches state-of-the-art levels of 91.81\% and 93.23\% for ``presence'' and ``comparison'' questions respectively, and improves from 90.00\% to 93.00\% on the ``Rural/Urban'' question. The improvement of UniRS over its base model, VILA-1.5~\cite{lin2024vila}, is particularly significant, improving from 66.44\% to 92.63\% on the average accuracy. 

\subsubsection{Comparision on RSVQA-HR} We further evaluate UniRS on the RSVQA-HR~\cite{lobry2020rsvqa} dataset under zero-shot setting, comparing it with general VLMs i.e., VILA1.5 (3B)~\cite{lin2024vila}, MiniGPTv2~\cite{chen2023minigpt} and LLaVA-1.5~\cite{liu2024visual}, as well as remote sensing-specific VLMs i.e., GeoChat~\cite{kuckreja2024geochat} and EarthGPT~\cite{zhang2024earthgpt}. As shown in TABLE~\ref{tab:rsvqa-hrben}, UniRS demonstrates superior remote sensing understanding capabilities in the high-resolution visual question answering task under zero-shot conditions. UniRS achieves an accuracy of 84.05\% on ``comparison'' questions and an average accuracy of 73.15\%, surpassing GeoChat and establishing a new state-of-the-art. 

\subsubsection{Comparision on CRSVQA} Additionally, we test UniRS's remote sensing analysis and instruction-following capabilities on the CRSVQA~\cite{zhang2023multi} test set under a supervised setting, with results presented in TABLE~\ref{tab:crsvqa}. During supervised training, we introduce the prompt augmentation mechanism to the instruction. UniRS is compared against the base model, traditional expert models and remote sensing VLMs, i.e., EarthGPT~\cite{zhang2024earthgpt} and GeoChat~\cite{kuckreja2024geochat}. The results show that UniRS significantly outperforms both traditional expert models and the other VLMs on the CRSVQA dataset. For instance, UniRS achieves an overall accuracy of 86.67\%, which is significantly higher than GeoChat's 82.50\% and improves over the best-performing traditional expert model, MQVQA~\cite{zhang2023multi}, by 15.76\%. Furthermore, it outperforms the base model, VILA-1.5~\cite{lin2024vila} by 6.34\%, showing the efficiency of our method.

The above quantitative results validate the strong generalization capability of UniRS in remote sensing visual question answering tasks. Compared to its base model, instruction fine-tuning notably enhances the multimodal remote sensing analysis capability of the VLM. Furthermore, UniRS, built on LLM foundation, demonstrates superior performance compared to traditional expert models.

\begin{figure*}[h]
    \centering
    \subfloat[]{%
        \includegraphics[width=0.6\linewidth]{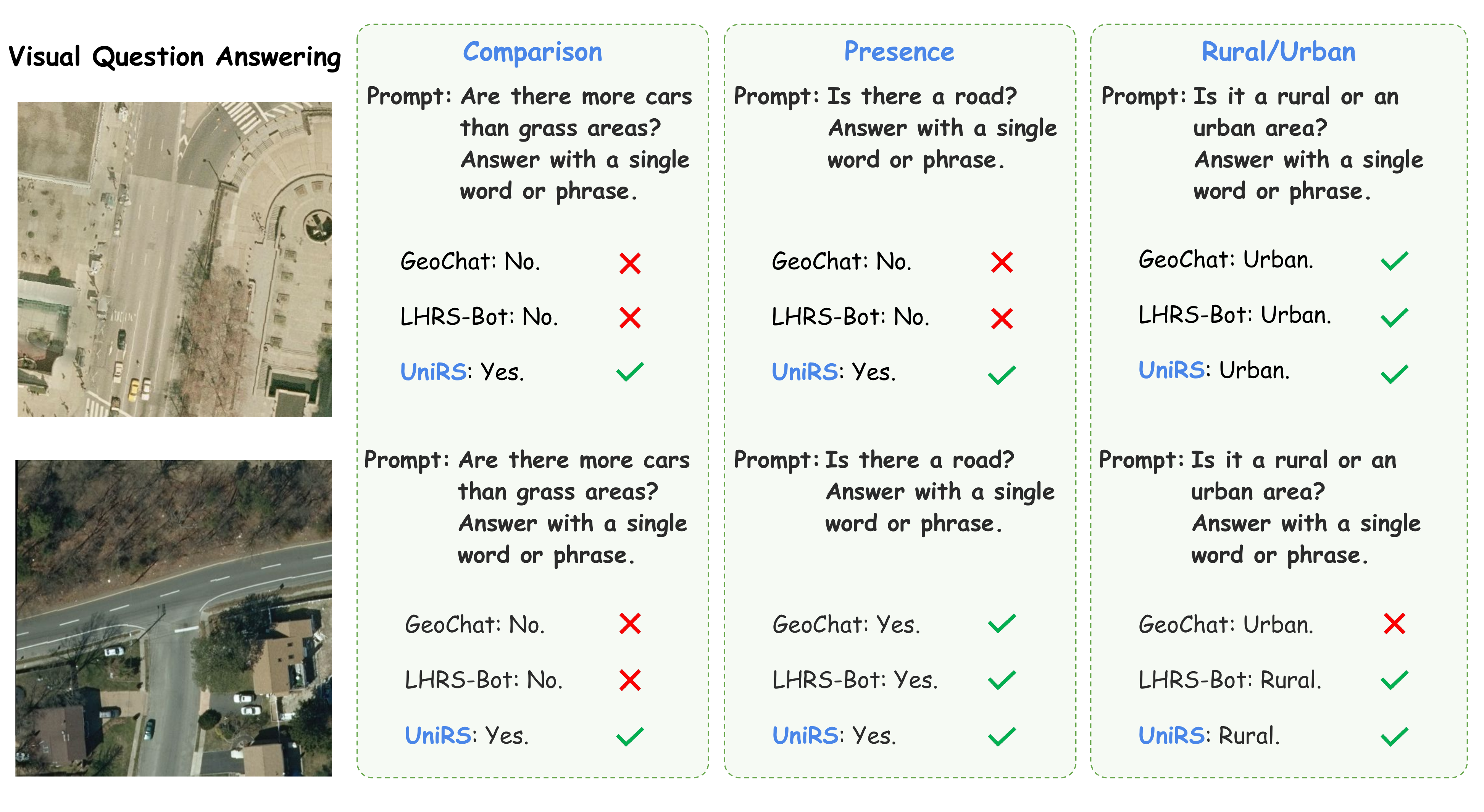}%
        \label{fig:sub1}}
    \hfill
    \subfloat[]{%
        \includegraphics[width=0.4\linewidth]{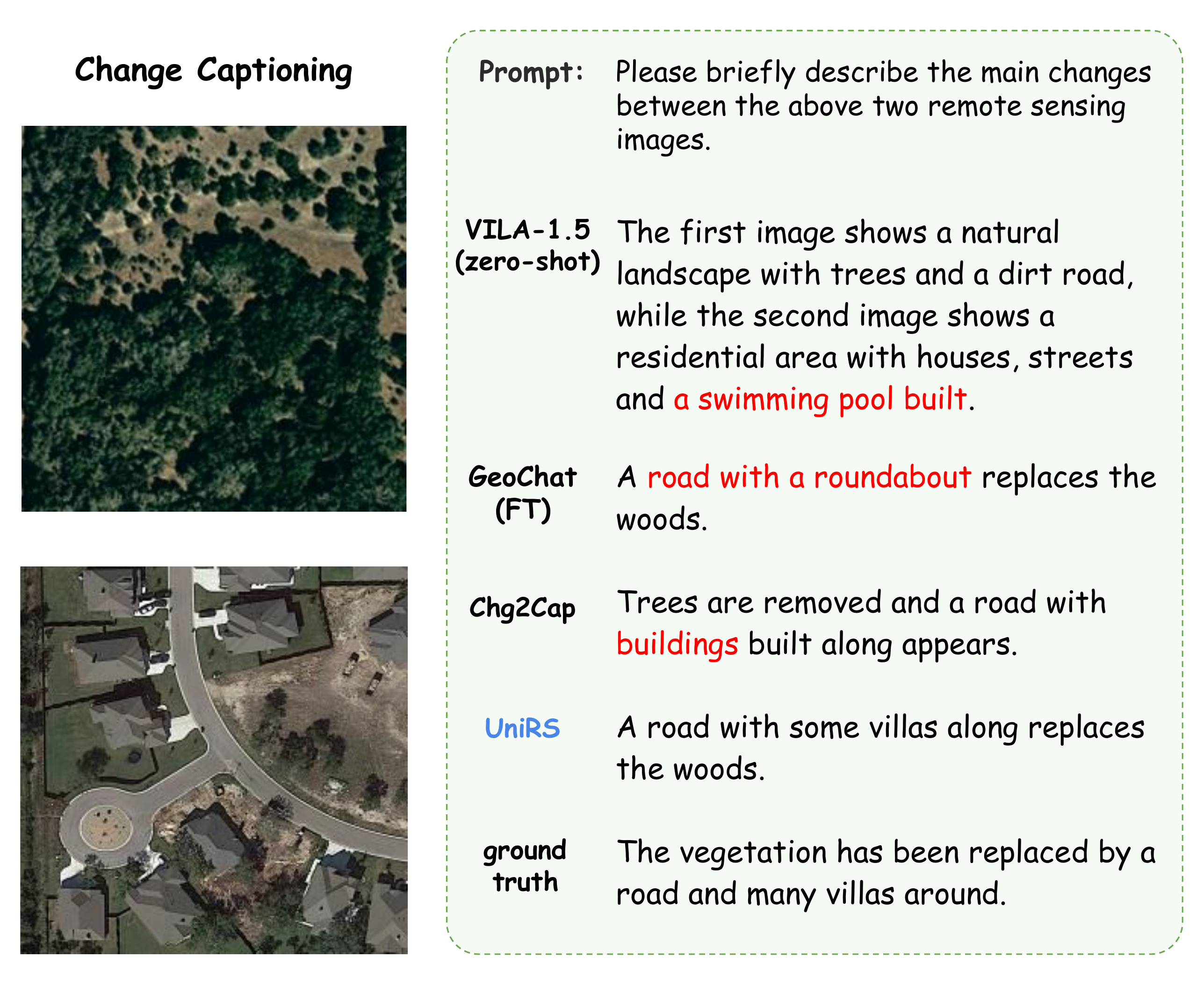}%
        \label{fig:sub2}}
    \caption{Qualitative results of our UniRS on visual question answering~\ref{fig:sub1} and change captioning~\ref{fig:sub2}. We compare our UniRS with other remote sensing VLMs on samples randomly selected. The incorrect responses are highlighted in \red{red}.}
    \label{fig:two_subfigures}
    \vspace{-0.35cm}
\end{figure*}

\textbf{Qualitative Results.}
We select two different remote sensing images and ask UniRS and other remote sensing VLMs the same questions for qualitative analysis, covering the categories of ``presence’’, ``comparison’’, and ``Rural/Urban’’. The results, shown in Fig~\ref{fig:sub1}, demonstrate that UniRS performs satisfactorily across remote sensing tasks with varying levels of granularity, outperforming other mainstream remote sensing VLMs, i.e., GeoChat~\cite{kuckreja2024geochat} and LHRS-Bot~\cite{muhtar2024lhrs}. The most notable improvement is in ``comparison’’ questions, where UniRS is able to finely capture the quantitative relationships of objects in the image and make correct comparisons, while the other two VLMs lack the precision required for this task. In the second question (top center), the other two models fail to correctly identify the road, possibly influenced by the overall image tone, while UniRS makes the correct judgment. In the overall scene classification, GeoChat incorrectly classifies the second image as 'urban', while UniRS and LHRS-Bot make the correct classification. Our UniRS may benefit from the prompt augmentation mechanism, which provides UniRS with the clues for reasoning, and the joint fine-tuning, which enhances the model’s spatial feature understanding.

\subsection{Remote Sensing Change Captioning}

We first introduce the datasets used for testing, followed by a quantitative comparison of UniRS with different methods. Then, a qualitative analysis highlights UniRS's performance advantage over other methods in the change captioning task.

\textbf{Dataset.} We test UniRS's understanding of dual-time image pair inputs on the LEVIR-CC test set. The test set contains 1,929 pairs of dual-time images. Among these, 964 samples are labeled as having changes of interest, while 965 samples are labeled negative. The captions use concise language to describe the significant objects of change and the specific changes within the image pairs.

\textbf{Quantitative Results.} We compare UniRS with several models on the LEVIR-CC~\cite{liu2022remote} dataset using the CIDEr-D metric, including the zero-shot base model, traditional expert models (i.e., Chg2Cap~\cite{chang2023changes}, PromptCC~\cite{liu2023decoupling}, PSNet~\cite{liu2023progressive}, and RSICCFormer~\cite{liu2022remote}), and VLMs fine-tuned on the training set. LLaVA-1.5~\cite{liu2024visual} and GeoChat~\cite{kuckreja2024geochat}, through engineering modifications, concatenate image features encoded from dual-time image pairs, input them into the LLM and get fine-tuning on the LEVIR-CC training set for evaluation. The experimental results, shown in TABLE~\ref{tab:levir_cc}, demonstrate that UniRS outperforms all previous expert models and VLMs, achieving state-of-the-art performance on this task. UniRS achieves a CIDEr-D score of 139.12, which is 2.51 higher than the previous SOTA, Chg2Cap~\cite{chang2023changes}, and 10.76 higher than GeoChat, another remote sensing VLM. This indicates that our designed change extraction module for dual-time image pair input effectively captures rich spatiotemporal features, aiding the model in performing the change captioning task. In contrast, the base model, VILA-1.5~\cite{lin2024vila}, only achieves a CIDEr-D score of 6.22 in the zero-shot setting, strongly validating the significant effectiveness of our method.
% We compare UniRS with two categories of models on the LEVIR-CC dataset with CIDEr-D metric: transformer decoder network based i.e., Chg2Cap~\cite{chang2023changes}, PSNet~\cite{liu2023progressive} and RSICCFormer~\cite{liu2022remote}; LLM based i.e., PromptCC~\cite{liu2023decoupling}, LLaVA-1.5~\cite{liu2024visual} and GeoChat~\cite{kuckreja2024geochat}. Among these, PromptCC is not an instruction-following model, while LLaVA-1.5 and GeoChat are engineered to support dual image input and fine-tuned for 5 epochs on the LEVIR-CC training set before testing. As shown in TABLE~\ref{tab:levir_cc}, UniRS achieves a CIDEr-D score of 139.12, outperforming Chg2Cap by 2.51 and reaching the new SOTA. In contrast, LLaVA-1.5 and GeoChat score 126.25 and 128.36 respectively after tuning while the VILA-1.5 (3B) only scores 6.22 in the zero-shot setting, demonstrating UniRS's strong generalization ability in dual-image change caption task and the effectiveness of our design.

\textbf{Qualitative Results.} We conduct a qualitative experiment to compare the base model (i.e. VILA-1.5 (3B)~\cite{lin2024vila}), fine-tuned GeoChat~\cite{kuckreja2024geochat}, previous SOTA method (i.e. Chg2Cap~\cite{chang2023changes}) and UniRS. As shown in Fig.~\ref{fig:sub2}, UniRS accurately understands the objects and change features of interest, providing concise descriptions. Although the VILA-1.5 model can comprehend the spatiotemporal features of the image pair, it produces a low-density description and suffers from hallucination, mentioning a non-existent swimming pool. Additionally, the fine-tuned GeoChat fails to capture the presence of houses along the road, demonstrating an insufficient understanding of the spatiotemporal features. Moreover, Chg2Cap's wording is inaccurate, describing the residential villas as buildings. This analysis shows that UniRS can effectively extract spatiotemporal changes of interest and demonstrates the generalization ability of UniRS in the change captioning task.

\begin{table*}[tb]
	\begin{center}
        \caption{Comparison of video scene classification performance on ERA dataset~\cite{mou2020era}. Here HDense~\cite{csahin2023deep}, FuTH-Net~\cite{jin2022futh}, MSTN~\cite{yang2022multiscale}, TRM~\cite{jin2021temporal} and ASAT~\cite{shi2023alleviating} are expert models designed for video classification. VILA-1.5 (3B) is tested under the zero-shot setting.}
		\label{table:era_video}
        \renewcommand{\arraystretch}{1.2}\resizebox{\linewidth}{!}{
			\begin{tabular}{c|ccccccccccccccccccccccccc|c}
				\hline
                \textbf{Method} &
  \textbf{\rotatebox{70}{post-earthquake}} &
  \textbf{\rotatebox{70}{flood}} &
  \textbf{\rotatebox{70}{fire}} &
  \textbf{\rotatebox{70}{landslide}} &
  \textbf{\rotatebox{70}{mudslide}} &
  \textbf{\rotatebox{70}{traffic collision}} &
  \textbf{\rotatebox{70}{traffic congestion}} &
  \textbf{\rotatebox{70}{harvesting}} &
  \textbf{\rotatebox{70}{ploughing}} &
  \textbf{\rotatebox{70}{constructing}} &
  \textbf{\rotatebox{70}{police chase}} &
  \textbf{\rotatebox{70}{conflict}} &
  \textbf{\rotatebox{70}{baseball}} &
  \textbf{\rotatebox{70}{basketball}} &
  \textbf{\rotatebox{70}{boating}} &
  \textbf{\rotatebox{70}{cycling}} &
  \textbf{\rotatebox{70}{running}} &
  \textbf{\rotatebox{70}{soccer}} &
  \textbf{\rotatebox{70}{swimming}} &
  \textbf{\rotatebox{70}{car racing}} &
  \textbf{\rotatebox{70}{party}} &
  \textbf{\rotatebox{70}{concert}} &
  \textbf{\rotatebox{70}{parade/protest}} &
  \textbf{\rotatebox{70}{religious activity}} &
  \textbf{\rotatebox{70}{non-event}} &
  \textbf{OA} \\
  \hline
  VILA-1.5 (3B)~\cite{lin2024vila} &
  0.0 & 63.3 & 14.3 & 16.3 & 0.0 & 35.8 & 70.0 & 28.1 & 34.6 & 40.7 & 0.0 & \underline{68.0} & \underline{86.0} &
  58.3 & 62.7 & 26.4 & 12.8 & 85.5 & 11.8 & \textbf{100.0} & 0.0 & \underline{91.8} & 55.1 & 5.6 &
  \underline{78.1} & 41.7 \\ \hline
  HDense~\cite{csahin2023deep} &
  67.3 & 71.4 & 78.6 & 34.7 & 74.5 & 35.9 & \underline{74.0} & \underline{81.3} & 82.7 & 59.3 & 64.7 & 16.0 & 76.0 & 72.9 & \underline{88.2} & 62.3 & 16.3 & 82.3 & 76.5 & 63.2 & 54.0 & 73.5 & 59.2 & 61.1 & 58.1 & 63.0 \\
FuTH-Net~\cite{jin2022futh} &
  \underline{72.7} & 75.7 & 87.5 & 57.1 & 74.5 & 34.0 & 56.0 & 76.6 & 71.2 & 81.4 & 76.5 & 36.0 & 78.0 &
  85.4 & 80.4 & 73.6 & 16.3 & 64.5 & 80.4 & 84.2 & 56.0 & 89.8 & \underline{65.3} & 63.0 & 63.9 & 66.8 \\
  MSTN~\cite{yang2022multiscale} & 61.8 & 76.1 & \underline{92.2} & \underline{60.4} & 62.8 & \underline{54.1} & 69.6 & 80.0 & \underline{91.1} & 73.6 & 71.7 & 54.6 & 86.0 & 72.4 & 86.5 & 66.0 & 66.9 & 90.2 & 74.1 & 61.9 & 67.4 & 56.0 & 46.6 & 58.5 & 51.5 & 67.4 \\ 
  TRM~\cite{jin2021temporal} & 72.7 & 75.5 & 87.5 & 57.1 & \underline{74.5} & 34.0 & 56.0 & 76.6 & 71.2 & \underline{81.4} & 76.5 & 36.0 & 78.0 & \underline{85.4} & 80.4 & \underline{73.6} & 16.3 & 64.5 & \underline{80.4} & 84.2 & 56.0 & 89.8 & 65.3 & 63.0 & 63.9 & 66.8 \\
  ASAT~\cite{shi2023alleviating} &
  62.3 & \underline{85.7} & 91.4 & 56.5 & 62.7 & 47.7 & 66.0 & 68.8 & 90.8 & 75.0 & \underline{80.5} & 40.0 & 84.0 & 78.9 & 85.7 & 64.6 & \textbf{78.8} & \textbf{94.0} & 61.4 & 61.9 & \textbf{87.1} & 56.0 & 47.9 & \underline{65.3} & 55.2 & \underline{68.1} \\
  \rowcolor{blue!15}
UniRS (ours) &
  \textbf{85.5} &
  \textbf{89.8} &
  \textbf{100.0} &
  \textbf{69.4} &
  \textbf{78.4} &
  \textbf{67.9} &
  \textbf{88.0} &
  \textbf{95.3} &
  \textbf{94.2} &
  \textbf{91.5} &
  \textbf{88.2} &
  \textbf{96.0} &
  \textbf{96.0} &
  \textbf{100.0} &
  \textbf{100.0} &
  \textbf{94.3} &
  \underline{73.3} &
  \underline{91.9} &
  \textbf{88.2} &
  \underline{89.5} &
  \underline{86.0} &
  \textbf{93.9} &
  \textbf{83.7} &
  \textbf{92.6} &
  \textbf{82.6} &
  \textbf{87.8} \\ \hline
\end{tabular}}
	\end{center}
\vspace{-0.35cm}
\end{table*}

\begin{figure}[tb]
  \centering
   \includegraphics[width=0.85\linewidth]{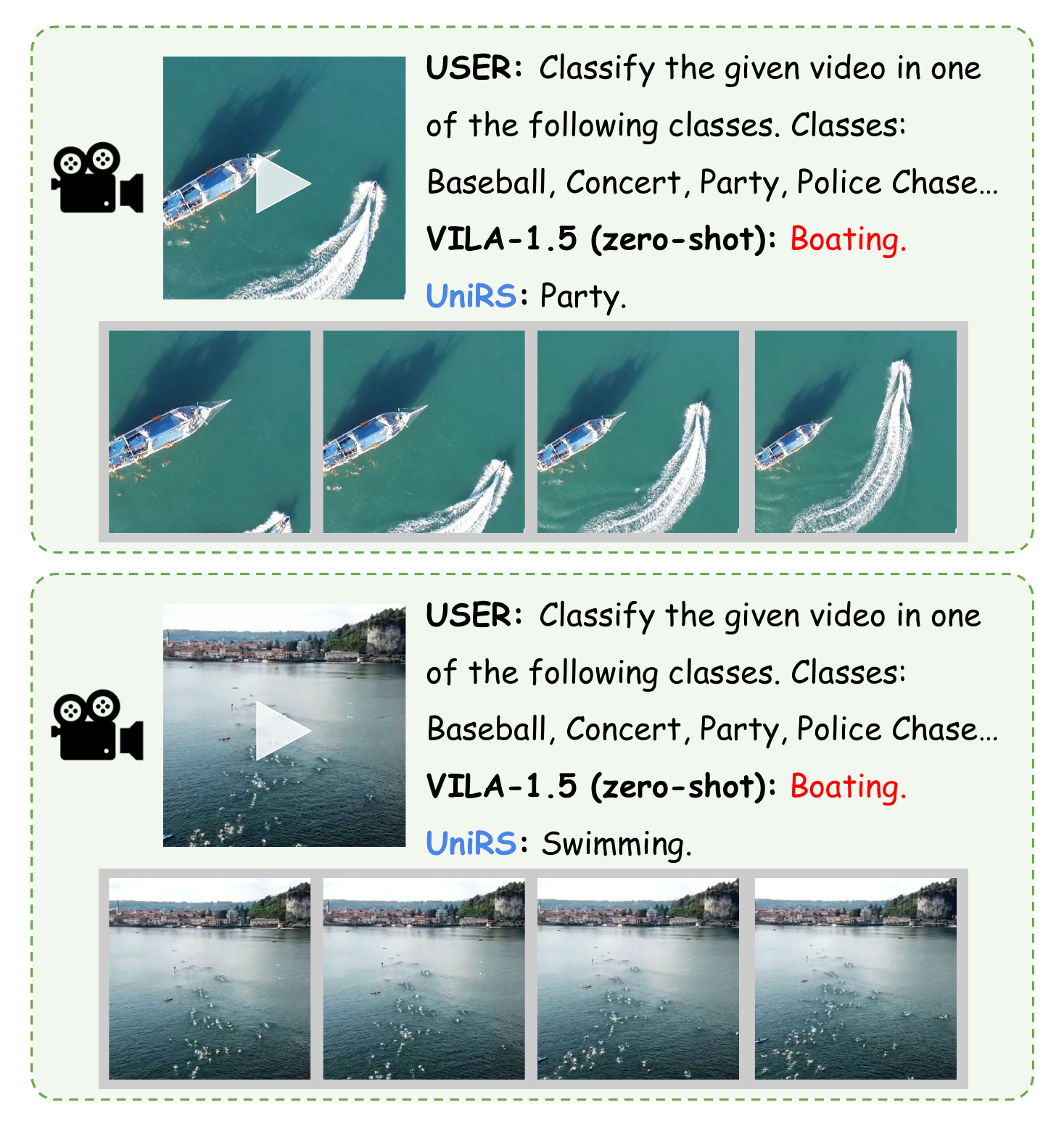}

   \caption{
   Qualitative results of our UniRS on video scene classification.
   }
   \label{fig:Qualtative_ERA}
\vspace{-0.35cm}
\end{figure}

\subsection{Remote Sensing Video Classification}

We sequentially introduce the datasets used for testing, the quantitative results, and the qualitative analysis, where we compare UniRS with the base model VILA-1.5~\cite{lin2024vila} to demonstrate the effectiveness of our work.

\textbf{Dataset.} The evaluation of remote sensing video understanding capability is conducted on the test set of the ERA Dataset, which contains 1,391 annotated videos, categorized into 25 common event classes. These videos, a significant type of data in remote sensing video, are captured from an overhead perspective by UAV. During the evaluation process, we record two metrics: Overall Accuracy(OA) and Per-class Precision.

\begin{table*}[tbh]
	\begin{center}
\caption{
  Ablation study on joint training. Here, ``individual'' refers to individual training of the model on the GeoChat-Instruct~\cite{kuckreja2024geochat}, LEVIR-CC~\cite{liu2022remote}, and ERA Dataset~\cite{mou2020era}. The performance is evaluated on three tasks: Visual Question Answering (i.e., RSVQA-LR, RSVQA-HR, and CRSVQA)~\cite{lobry2020rsvqa, zhang2023multi}, Change Captioning~\cite{liu2022remote}, and Video Scene Classification~\cite{mou2020era}.
  }
		\label{table:ablation1}
        \renewcommand{\arraystretch}{1.2}\resizebox{\linewidth}{!}{
			\begin{tabular}{c|cccc|ccc|c|c|c}
\hline
\multirow{2}{*}{\textbf{Method}} &
  \multicolumn{4}{c|}{\textbf{RSVQA-LR}} &
  \multicolumn{3}{c|}{\textbf{RSVQA-HR}} &       \textbf{CRSVQA} &
  \textbf{Change Captioning} &
  \textbf{Video Scene Classification} \\ \cline{2-11} 
 &
  \textbf{Presence} &
  \textbf{Comparison} &
  \textbf{Rural/Urban} &
  \textbf{Avg. Accuracy} &
  \multicolumn{1}{c}{\textbf{Presence}} &
  \textbf{Comparison} &
  \textbf{Avg. Accuracy} &       \textbf{Overall Accuracy} &
  \textbf{CIDEr-D} &
  \textbf{Overall Accuracy} \\ \hline
UniRS (individual) & 91.51 & 91.38 & \textbf{92.00} & 91.44 & \textbf{61.07} & 80.03 & 71.68 & 83.33 &  137.61    &  84.4    \\ \hline \rowcolor{blue!15}
UniRS (ours) &
  \textbf{91.64} &
  \textbf{92.68} &
  90.00 &
  \textbf{92.21} &
  59.29 &
  \textbf{84.05} &
  \textbf{73.15} &
  \textbf{86.67} &
  \textbf{139.12} &
  \textbf{87.8} \\ \hline
\end{tabular}}
	\end{center}
\vspace{-0.35cm}
\end{table*}

\textbf{Quantitative Results.} We compare our UniRS with expert models(i.e., HDense~\cite{csahin2023deep}, FuTH-Net~\cite{jin2022futh}, MSTN~\cite{yang2022multiscale}, TRM~\cite{jin2021temporal}, and ASAT~\cite{shi2023alleviating}) using per-class precision and overall accuracy metrics, with results shown in TABLE~\ref{table:era_video}. It is evident that UniRS has significant advantages in remote sensing video understanding. The previous best expert model, ASAT, achieves an overall accuracy of 68.1\%, while UniRS reaches 87.8\%, outperforming ASAT by 19.7\% and achieving the new SOTA. In terms of per-class precision, UniRS demonstrates outstanding performance in all scene categories. Although it does not reach SOTA in the categories of ``running'', ``soccer'', ``car racing'' and ``party'', UniRS still exhibits exceptional levels of performance in these classes. Additionally, VILA-1.5 (3B) shows significant deficiencies in zero-shot understanding of certain events, achieving a precision of 0.0 in the ``party'' category and 100.0 in the ``car racing'' category. However, through joint fine-tuning, these shortcomings are effectively compensated.

\textbf{Qualitative Results.} We conduct a qualitative analysis of video scene classification task, comparing the performance of UniRS with its base model, VILA-1.5~\cite{lin2024vila}. The results in Fig.~\ref{fig:Qualtative_ERA} reveal that UniRS exhibits a stronger ability to understand both global and local spatiotemporal features. In the first video, which is somewhat misleading, VILA-1.5 focuses on the speedboat's movement and overlooks the activities of the passengers and the surroundings, misclassifying the scene as ``boating’’. In contrast, UniRS makes the correct judgment. In the second video, which features a small boat and many swimmers making a splash on the river, VILA-1.5 fails to correctly identify the majority of swimmers in the scene, misclassifying it as ``boating’’, while UniRS provides the correct answer. This demonstrates that our joint fine-tuning on multi-temporal tasks enhances the model's ability to understand the spatiotemporal information embedded in videos.
% We select video samples from the party and car racing categories for qualitative analysis. VILA-1.5 (3B) achieved a precision of 0.0 in the ``party'' category and 100.0 in the ``car racing'' category. As shown in Fig.~\ref{fig:Qualtative_ERA}, the base model, VILA-1.5 (3B), demonstrates a basic understanding of remote sensing aerial-view scene videos under zero-shot setting, while our UniRS can focus on temporal and spatial features of interest. VILA incorrectly classifies our selected party video as the ``boating'' category. This video presents certain challenges, as it depicts a yacht party at sea, prominently featuring a boat in motion, which may mislead the model to categorize it as ``boating''. Accurate scene recognition requires the model to pay closer attention to the behaviors of people aboard and around the boat. In contrast, for the distinctly featured car racing videos, both UniRS and VILA correctly classify the content. This indicates that our method helps the model focus on meaningful temporal and local spatial features in the videos, thereby enhancing its video understanding capabilities.

\subsection{Ablation Studies}

\begin{table}[t]
\centering
\caption{
Ablation study on change extraction module on LEVIR-CC dataset~\cite{liu2022remote}. We study the influence of the Change Extraction module on the performance with and without joint training.
}
\label{tab:ablation_2}
\renewcommand{\arraystretch}{1.1} \resizebox{0.35\textwidth}{!}{
\begin{tabular}{cc|c}
\hline
\multicolumn{2}{c|}{\textbf{Modules}}               & \multicolumn{1}{c}{\multirow{2}{*}{\textbf{CIDEr-D}}} \\ \cline{1-2}
\textbf{Joint Training} & \textbf{Change Extraction Module} & \multicolumn{1}{c}{}                         \\ \hline
\rowcolor{gray!15}
               &                          & 126.41                                       \\
\Checkmark              &                          & 131.28                                       \\ \hline
               & \Checkmark                        & 137.61                                       \\        \rowcolor{blue!15}
\Checkmark              & \Checkmark                        & \textbf{139.12}                              \\ \hline
\end{tabular}}
\vspace{-0.35cm}
\end{table}

\begin{table*}[tbh]
\begin{center}
\caption{
Ablation study on prompt augmentation mechanism (PA). The performance is evaluated on three tasks: visual question answering (i.e., RSVQA-LR, RSVQA-HR, and CRSVQA)~\cite{lobry2020rsvqa, zhang2023multi}, change captioning~\cite{liu2022remote} and video scene classification~\cite{mou2020era}.
}
\label{tab:ablation_3}
\renewcommand{\arraystretch}{1.2} \resizebox{1.0\textwidth}{!}{
\begin{tabular}{c|cccc|ccc|c|c|c}
\hline
\multirow{2}{*}{\textbf{Method}} &
  \multicolumn{4}{c|}{\textbf{RSVQA-LR}} &
  \multicolumn{3}{c|}{\textbf{RSVQA-HR}} &       \textbf{CRSVQA} &
  \textbf{Change Captioning} &
  \textbf{Video Scene Classification} \\ \cline{2-11} 
 &
  \textbf{Presence} &
  \textbf{Comparison} &
  \textbf{Rural/Urban} &
  \textbf{Avg. Accuracy} &
  \multicolumn{1}{c}{\textbf{Presence}} &
  \textbf{Comparison} &
  \textbf{Avg. Accuracy} &       \textbf{Overall Accuracy} &
  \textbf{CIDEr-D} &
  \textbf{Overall Accuracy} \\ \hline
UniRS (without PA) & \textbf{91.68} & 91.85 & \textbf{92.00} & 91.80 & 56.95 & 83.91 & 72.21 & 84.00 &  138.47    &  86.1    \\ \hline \rowcolor{blue!15}
UniRS (ours) &
  91.64 &
  \textbf{92.68} &
  90.00 &
  \textbf{92.21} &
  \textbf{59.29} &
  \textbf{84.05} &
  \textbf{73.15} &
  \textbf{86.67} &
  \textbf{139.12} &
  \textbf{87.8} \\ \hline
\end{tabular}}
\end{center}
\vspace{-0.35cm}
\end{table*}

\subsubsection{Ablation Study on Joint Instruction-tuning.} This section investigates the effectiveness of joint instruction-tuning across multi-temporal tasks of three types of visual inputs, i.e., single image, dual-time image pair, and video. During the training of UniRS, we perform joint training using a mix of GeoChat-Instruct~\cite{kuckreja2024geochat}, LEVIR-CC~\cite{liu2022remote}, and ERA~\cite{mou2020era} datasets. In the control group, we train models separately on each of these three datasets using the same initialization method as UniRS. We test the model fine-tuned with GeoChat-Instruct on the RSVQA-LR, -HR, and CRSVQA~\cite{zhang2023multi} test sets. The model trained on LEVIR-CC is evaluated on the LEVIR-CC test set, and the model trained individually on ERA is tested on the video scene classification task using the ERA test set.

The experimental results are shown in TABLE~\ref{table:ablation1}. Except for a modest improvement in the RSVQA-LR dataset compared to the control group, UniRS exhibits significant improvements in the other datasets under joint training. The average accuracy for RSVQA-HR increases from 71.68\% to 73.15\%. The CIDEr-D score on the LEVIR-CC test set improved from 137.61 to 139.12. In the ERA test set, the overall accuracy increased from 84.4\% to 87.8\%. These results demonstrate that under joint training, effective knowledge sharing across tasks has happened, validating the effectiveness of joint instruction fine-tuning.
% We compare the performance of models trained jointly and separately on corresponding multimodal tasks. As shown in Table~\ref{table:ablation1}, except for the test results on the LRBEN dataset, all other tests show significant improvements under joint training. Specifically, the Average Accuracy on the HRBEN dataset increases from 71.68\% with separate training to 74.38\%, the CIDEr-D score for the Change Captioning task improves from 137.61 to 139.17, and in the Video Classification task, the Overall Accuracy rises from 84.4\% to 87.5\%. These results indicate that joint training effectively facilitates knowledge sharing, enhancing the model's understanding across various visual input modalities.

\subsubsection{Ablation Study on Change Extraction Module.} This section validates the effectiveness of the Change Extraction module on the change captioning task. In the control group where the Change Extraction module is not used, we adopt a processing method similar to video. The visual features extracted from the dual-time image pair using the visual encoder are concatenated in temporal order and separated by a ``\textbackslash n'' token as input to the model. Additionally, control experiments are conducted for both joint training and individual training on the LEVIR-CC~\cite{liu2022remote} dataset to comprehensively assess the effectiveness of the module in different training scenarios. All experimental groups used the same initialization method, and the CIDEr-D score is reported to evaluate the results.

As reported in TABLE~\ref{tab:ablation_2}, the Change Extraction module demonstrates significant improvements in the change captioning task in both joint and individual training scenarios. When trained separately on the LEVIR-CC dataset without the Change Extraction module, the model achieves a CIDEr-D of 126.41. After incorporating the module, it increases to 137.61. In the joint training scenario, the inclusion of the Change Extraction module leads to an improvement in the CIDEr-D from 131.28 to 139.12. These results indicate that the Change Extraction module's enhancement of spatiotemporal relational information can effectively improve the model's ability to understand the visual features of dual-time image pairs.

% We also conduct an ablation study on the Change Extraction Module, which is used exclusively during the inference process when processing dual-time image pairs. This experiment is carried out on the change captioning task, with results presented in Table~\ref{tab:ablation_2}. When training on the LEVIR-CC dataset without the Change Extraction Module, the image features are projected into two visual embeddings and fed into the LLM, resulting in a CIDEr-D of 126.41. With the module applied, it increases to 137.61. Under joint training, the CIDEr-D further improves from 137.61 to 140.82. These results demonstrate the effectiveness of the Change Extraction Module in capturing spatiotemporal relationships within image pairs.

\subsubsection{Ablation Study on Prompt Augmentation Mechanism.} In this section, we study the importance and effectiveness of the prompt augmentation mechanism on three multi-temporal tasks, i.e., visual question answering, change captioning, and video scene classification. In the control group, we directly input the instructions from the datasets into the model during training and testing. The control group uses the same initialization settings as UniRS. After joint training on the same mixed dataset, comparative tests are conducted on each task.

As shown in TABLE~\ref{tab:ablation_3}, the prompt augmentation mechanism significantly improves the model's performance across all tasks, especially for high-resolution inputs. In the RSVQA-HR~\cite{lobry2020rsvqa}, the prompt augmentation mechanism increases the accuracy for ``Presence'' questions from 56.95\% to 59.29\%, and the overall average accuracy rises from 72.21\% to 73.15\%. In the CRSVQA~\cite{zhang2023multi} dataset, the overall accuracy improves from 84.00\% to 86.67\%, while in the ERA~\cite{mou2020era} test set, the overall accuracy increases from 86.1\% to 87.8\%. However, the improvements are modest in the RSVQA-LR~\cite{lobry2020rsvqa} and LEVIR-CC~\cite{liu2022remote} test sets, likely due to the lower image resolution of 256$\times$256, which reduces the reliability of the clues provided by the prompt augmentation mechanism. In contrast, high-resolution remote sensing images enable the prompt augmentation mechanism to offer more effective feature cues.

\section{Conclusion}

 Research about Vision Language Models~(VLMs) in remote sensing has made significant progress though, current models are limited to a narrow range of multimodal tasks and overlook the richness of visual information that needs analyzing in remote sensing. To address this gap, we propose UniRS, the first VLM to unify diverse multi-temporal remote sensing tasks of various input types: single image, dual-time image pair, and video, further exploring the representational potential of VLMs in remote sensing. We employ a unified, task-guided visual representation for the three types of tasks, design a VLM-based prompt augmentation mechanism hinting inference process, and develop a modality-specific spatiotemporal relationship extraction module for tasks with dual-time image pair input to help the model better understand fine-grained visual features. After joint instruction fine-tuning on a mixed dataset, UniRS learns from various types of visual inputs and demonstrates superior performance in visual question answering, change captioning, and video scene classification.

\bibliographystyle{IEEEtran}
\bibliography{reference}

\begin{thebibliography}{10}
\providecommand{\url}[1]{#1}
\csname url@samestyle\endcsname
\providecommand{\newblock}{\relax}
\providecommand{\bibinfo}[2]{#2}
\providecommand{\BIBentrySTDinterwordspacing}{\spaceskip=0pt\relax}
\providecommand{\BIBentryALTinterwordstretchfactor}{4}
\providecommand{\BIBentryALTinterwordspacing}{\spaceskip=\fontdimen2\font plus
\BIBentryALTinterwordstretchfactor\fontdimen3\font minus \fontdimen4\font\relax}
\providecommand{\BIBforeignlanguage}[2]{{%
\expandafter\ifx\csname l@#1\endcsname\relax
\typeout{** WARNING: IEEEtran.bst: No hyphenation pattern has been}%
\typeout{** loaded for the language `#1'. Using the pattern for}%
\typeout{** the default language instead.}%
\else
\language=\csname l@#1\endcsname
\fi
#2}}
\providecommand{\BIBdecl}{\relax}
\BIBdecl

\bibitem{liu2024visual}
H.~Liu, C.~Li, Q.~Wu, and Y.~J. Lee, ``Visual instruction tuning,'' \emph{Advances in neural information processing systems}, vol.~36, 2024.

\bibitem{alayrac2022flamingo}
J.-B. Alayrac, J.~Donahue, P.~Luc, A.~Miech, I.~Barr, Y.~Hasson, K.~Lenc, A.~Mensch, K.~Millican, M.~Reynolds \emph{et~al.}, ``Flamingo: a visual language model for few-shot learning,'' \emph{Advances in neural information processing systems}, vol.~35, pp. 23\,716--23\,736, 2022.

\bibitem{lin2024vila}
J.~Lin, H.~Yin, W.~Ping, P.~Molchanov, M.~Shoeybi, and S.~Han, ``Vila: On pre-training for visual language models,'' in \emph{Proceedings of the IEEE/CVF Conference on Computer Vision and Pattern Recognition}, 2024, pp. 26\,689--26\,699.

\bibitem{chen2023minigpt}
J.~Chen, D.~Zhu, X.~Shen, X.~Li, Z.~Liu, P.~Zhang, R.~Krishnamoorthi, V.~Chandra, Y.~Xiong, and M.~Elhoseiny, ``Minigpt-v2: large language model as a unified interface for vision-language multi-task learning,'' \emph{arXiv preprint arXiv:2310.09478}, 2023.

\bibitem{li2022blip}
J.~Li, D.~Li, C.~Xiong, and S.~Hoi, ``Blip: Bootstrapping language-image pre-training for unified vision-language understanding and generation,'' in \emph{International conference on machine learning}.\hskip 1em plus 0.5em minus 0.4em\relax PMLR, 2022, pp. 12\,888--12\,900.

\bibitem{li2023blip}
J.~Li, D.~Li, S.~Savarese, and S.~Hoi, ``Blip-2: Bootstrapping language-image pre-training with frozen image encoders and large language models,'' in \emph{International conference on machine learning}.\hskip 1em plus 0.5em minus 0.4em\relax PMLR, 2023, pp. 19\,730--19\,742.

\bibitem{touvron2023llama}
H.~Touvron, T.~Lavril, G.~Izacard, X.~Martinet, M.-A. Lachaux, T.~Lacroix, B.~Rozi{\`e}re, N.~Goyal, E.~Hambro, F.~Azhar \emph{et~al.}, ``Llama: Open and efficient foundation language models,'' \emph{arXiv preprint arXiv:2302.13971}, 2023.

\bibitem{ouyang2022training}
L.~Ouyang, J.~Wu, X.~Jiang, D.~Almeida, C.~Wainwright, P.~Mishkin, C.~Zhang, S.~Agarwal, K.~Slama, A.~Ray \emph{et~al.}, ``Training language models to follow instructions with human feedback,'' \emph{Advances in neural information processing systems}, vol.~35, pp. 27\,730--27\,744, 2022.

\bibitem{bai2023qwen}
J.~Bai, S.~Bai, Y.~Chu, Z.~Cui, K.~Dang, X.~Deng, Y.~Fan, W.~Ge, Y.~Han, F.~Huang \emph{et~al.}, ``Qwen technical report,'' \emph{arXiv preprint arXiv:2309.16609}, 2023.

\bibitem{brown2020language}
T.~Brown, B.~Mann, N.~Ryder, M.~Subbiah, J.~D. Kaplan, P.~Dhariwal, A.~Neelakantan, P.~Shyam, G.~Sastry, A.~Askell \emph{et~al.}, ``Language models are few-shot learners,'' \emph{Advances in neural information processing systems}, vol.~33, pp. 1877--1901, 2020.

\bibitem{schuhmann2022laion}
C.~Schuhmann, R.~Beaumont, R.~Vencu, C.~Gordon, R.~Wightman, M.~Cherti, T.~Coombes, A.~Katta, C.~Mullis, M.~Wortsman \emph{et~al.}, ``Laion-5b: An open large-scale dataset for training next generation image-text models,'' \emph{Advances in Neural Information Processing Systems}, vol.~35, pp. 25\,278--25\,294, 2022.

\bibitem{li2024llava}
C.~Li, C.~Wong, S.~Zhang, N.~Usuyama, H.~Liu, J.~Yang, T.~Naumann, H.~Poon, and J.~Gao, ``Llava-med: Training a large language-and-vision assistant for biomedicine in one day,'' \emph{Advances in Neural Information Processing Systems}, vol.~36, 2024.

\bibitem{chiang2024mobility}
H.-T.~L. Chiang, Z.~Xu, Z.~Fu, M.~G. Jacob, T.~Zhang, T.-W.~E. Lee, W.~Yu, C.~Schenck, D.~Rendleman, D.~Shah \emph{et~al.}, ``Mobility vla: Multimodal instruction navigation with long-context vlms and topological graphs,'' \emph{arXiv preprint arXiv:2407.07775}, 2024.

\bibitem{lobry2020rsvqa}
S.~Lobry, D.~Marcos, J.~Murray, and D.~Tuia, ``Rsvqa: Visual question answering for remote sensing data,'' \emph{IEEE Transactions on Geoscience and Remote Sensing}, vol.~58, no.~12, pp. 8555--8566, 2020.

\bibitem{kafle2016answer}
K.~Kafle and C.~Kanan, ``Answer-type prediction for visual question answering,'' in \emph{Proceedings of the IEEE conference on computer vision and pattern recognition}, 2016, pp. 4976--4984.

\bibitem{zhang2023multi}
M.~Zhang, F.~Chen, and B.~Li, ``Multi-step question-driven visual question answering for remote sensing,'' \emph{IEEE Transactions on Geoscience and Remote Sensing}, 2023.

\bibitem{bazi2022bi}
Y.~Bazi, M.~M. Al~Rahhal, M.~L. Mekhalfi, M.~A. Al~Zuair, and F.~Melgani, ``Bi-modal transformer-based approach for visual question answering in remote sensing imagery,'' \emph{IEEE Transactions on Geoscience and Remote Sensing}, vol.~60, pp. 1--11, 2022.

\bibitem{chang2023changes}
S.~Chang and P.~Ghamisi, ``Changes to captions: An attentive network for remote sensing change captioning,'' \emph{IEEE Transactions on Image Processing}, 2023.

\bibitem{liu2023progressive}
C.~Liu, J.~Yang, Z.~Qi, Z.~Zou, and Z.~Shi, ``Progressive scale-aware network for remote sensing image change captioning,'' in \emph{IGARSS 2023-2023 IEEE International Geoscience and Remote Sensing Symposium}.\hskip 1em plus 0.5em minus 0.4em\relax IEEE, 2023, pp. 6668--6671.

\bibitem{jin2022futh}
P.~Jin, L.~Mou, Y.~Hua, G.-S. Xia, and X.~X. Zhu, ``Futh-net: fusing temporal relations and holistic features for aerial video classification,'' \emph{IEEE Transactions on Geoscience and Remote Sensing}, vol.~60, pp. 1--13, 2022.

\bibitem{marino2019ok}
K.~Marino, M.~Rastegari, A.~Farhadi, and R.~Mottaghi, ``Ok-vqa: A visual question answering benchmark requiring external knowledge,'' in \emph{Proceedings of the IEEE/cvf conference on computer vision and pattern recognition}, 2019, pp. 3195--3204.

\bibitem{csahin2023deep}
A.~H. {\c{S}}ahin and H.~F. Ate{\c{s}}, ``Deep learning based event recognition in aerial imagery,'' in \emph{2023 8th International Conference on Computer Science and Engineering (UBMK)}.\hskip 1em plus 0.5em minus 0.4em\relax IEEE, 2023, pp. 426--431.

\bibitem{liu2023decoupling}
C.~Liu, R.~Zhao, J.~Chen, Z.~Qi, Z.~Zou, and Z.~Shi, ``A decoupling paradigm with prompt learning for remote sensing image change captioning,'' \emph{IEEE Transactions on Geoscience and Remote Sensing}, 2023.

\bibitem{muhtar2024lhrs}
D.~Muhtar, Z.~Li, F.~Gu, X.~Zhang, and P.~Xiao, ``Lhrs-bot: Empowering remote sensing with vgi-enhanced large multimodal language model,'' \emph{arXiv preprint arXiv:2402.02544}, 2024.

\bibitem{kuckreja2024geochat}
K.~Kuckreja, M.~S. Danish, M.~Naseer, A.~Das, S.~Khan, and F.~S. Khan, ``Geochat: Grounded large vision-language model for remote sensing,'' in \emph{Proceedings of the IEEE/CVF Conference on Computer Vision and Pattern Recognition}, 2024, pp. 27\,831--27\,840.

\bibitem{zhan2024skyeyegpt}
Y.~Zhan, Z.~Xiong, and Y.~Yuan, ``Skyeyegpt: Unifying remote sensing vision-language tasks via instruction tuning with large language model,'' \emph{arXiv preprint arXiv:2401.09712}, 2024.

\bibitem{irvin2024teochat}
J.~A. Irvin, E.~R. Liu, J.~C. Chen, I.~Dormoy, J.~Kim, S.~Khanna, Z.~Zheng, and S.~Ermon, ``Teochat: A large vision-language assistant for temporal earth observation data,'' \emph{arXiv preprint arXiv:2410.06234}, 2024.

\bibitem{bazi2024rs}
Y.~Bazi, L.~Bashmal, M.~M. Al~Rahhal, R.~Ricci, and F.~Melgani, ``Rs-llava: A large vision-language model for joint captioning and question answering in remote sensing imagery,'' \emph{Remote Sensing}, vol.~16, no.~9, p. 1477, 2024.

\bibitem{hu2023rsgpt}
Y.~Hu, J.~Yuan, C.~Wen, X.~Lu, and X.~Li, ``Rsgpt: A remote sensing vision language model and benchmark,'' \emph{arXiv preprint arXiv:2307.15266}, 2023.

\bibitem{zhang2024earthgpt}
W.~Zhang, M.~Cai, T.~Zhang, Y.~Zhuang, and X.~Mao, ``Earthgpt: A universal multi-modal large language model for multi-sensor image comprehension in remote sensing domain,'' \emph{IEEE Transactions on Geoscience and Remote Sensing}, 2024.

\bibitem{liu2022remote}
C.~Liu, R.~Zhao, H.~Chen, Z.~Zou, and Z.~Shi, ``Remote sensing image change captioning with dual-branch transformers: A new method and a large scale dataset,'' \emph{IEEE Transactions on Geoscience and Remote Sensing}, vol.~60, pp. 1--20, 2022.

\bibitem{mou2020era}
L.~Mou, Y.~Hua, P.~Jin, and X.~X. Zhu, ``Era: A data set and deep learning benchmark for event recognition in aerial videos [software and data sets],'' \emph{IEEE Geoscience and Remote Sensing Magazine}, vol.~8, no.~4, pp. 125--133, 2020.

\bibitem{achiam2023gpt}
J.~Achiam, S.~Adler, S.~Agarwal, L.~Ahmad, I.~Akkaya, F.~L. Aleman, D.~Almeida, J.~Altenschmidt, S.~Altman, S.~Anadkat \emph{et~al.}, ``Gpt-4 technical report,'' \emph{arXiv preprint arXiv:2303.08774}, 2023.

\bibitem{luo2024zero}
D.~Luo, J.~Huang, S.~Gong, H.~Jin, and Y.~Liu, ``Zero-shot video moment retrieval from frozen vision-language models,'' in \emph{Proceedings of the IEEE/CVF Winter Conference on Applications of Computer Vision}, 2024, pp. 5464--5473.

\bibitem{springstein2024visual}
M.~Springstein, S.~Schneider, J.~Rahnama, J.~Stalter, M.~Kristen, E.~M{\"u}ller-Budack, and R.~Ewerth, ``Visual narratives: Large-scale hierarchical classification of art-historical images,'' in \emph{Proceedings of the IEEE/CVF Winter Conference on Applications of Computer Vision}, 2024, pp. 7220--7230.

\bibitem{radford2021learning}
A.~Radford, J.~W. Kim, C.~Hallacy, A.~Ramesh, G.~Goh, S.~Agarwal, G.~Sastry, A.~Askell, P.~Mishkin, J.~Clark \emph{et~al.}, ``Learning transferable visual models from natural language supervision,'' in \emph{International conference on machine learning}.\hskip 1em plus 0.5em minus 0.4em\relax PMLR, 2021, pp. 8748--8763.

\bibitem{zhai2023sigmoid}
X.~Zhai, B.~Mustafa, A.~Kolesnikov, and L.~Beyer, ``Sigmoid loss for language image pre-training,'' in \emph{Proceedings of the IEEE/CVF International Conference on Computer Vision}, 2023, pp. 11\,975--11\,986.

\bibitem{liu2024improved}
H.~Liu, C.~Li, Y.~Li, and Y.~J. Lee, ``Improved baselines with visual instruction tuning,'' in \emph{Proceedings of the IEEE/CVF Conference on Computer Vision and Pattern Recognition}, 2024, pp. 26\,296--26\,306.

\bibitem{luo2024skysensegpt}
J.~Luo, Z.~Pang, Y.~Zhang, T.~Wang, L.~Wang, B.~Dang, J.~Lao, J.~Wang, J.~Chen, Y.~Tan \emph{et~al.}, ``Skysensegpt: A fine-grained instruction tuning dataset and model for remote sensing vision-language understanding,'' \emph{arXiv preprint arXiv:2406.10100}, 2024.

\bibitem{cheng2022nwpu}
Q.~Cheng, H.~Huang, Y.~Xu, Y.~Zhou, H.~Li, and Z.~Wang, ``Nwpu-captions dataset and mlca-net for remote sensing image captioning,'' \emph{IEEE Transactions on Geoscience and Remote Sensing}, vol.~60, pp. 1--19, 2022.

\bibitem{yang2022multiscale}
F.~Yang, J.~Zhang, Y.~Zhao, A.~Qin, and C.~Gao, ``Multiscale spatio-temporal network for aerial video event recognition,'' in \emph{IGARSS 2022-2022 IEEE International Geoscience and Remote Sensing Symposium}.\hskip 1em plus 0.5em minus 0.4em\relax IEEE, 2022, pp. 7835--7838.

\bibitem{shi2023alleviating}
G.~Shi, X.~Fu, C.~Cao, and Z.-J. Zha, ``Alleviating spatial misalignment and motion interference for uav-based video recognition,'' in \emph{Proceedings of the 31st ACM International Conference on Multimedia}, 2023, pp. 193--202.

\bibitem{caruna1993multitask}
R.~Caruna, ``Multitask learning: A knowledge-based source of inductive bias,'' in \emph{Machine learning: Proceedings of the tenth international conference}, 1993, pp. 41--48.

\bibitem{caruana1997multitask}
R.~Caruana, ``Multitask learning,'' \emph{Machine learning}, vol.~28, pp. 41--75, 1997.

\bibitem{jaradat2024multitask}
S.~Jaradat, R.~Nayak, A.~Paz, H.~I. Ashqar, and M.~Elhenawy, ``Multitask learning for crash analysis: A fine-tuned llm framework using twitter data,'' \emph{Smart Cities}, vol.~7, no.~5, pp. 2422--2465, 2024.

\bibitem{chen2023tigerbot}
Y.~Chen, W.~Cai, L.~Wu, X.~Li, Z.~Xin, and C.~Fu, ``Tigerbot: An open multilingual multitask llm,'' \emph{arXiv preprint arXiv:2312.08688}, 2023.

\bibitem{liu2024mftcoder}
B.~Liu, C.~Chen, Z.~Gong, C.~Liao, H.~Wang, Z.~Lei, M.~Liang, D.~Chen, M.~Shen, H.~Zhou \emph{et~al.}, ``Mftcoder: Boosting code llms with multitask fine-tuning,'' in \emph{Proceedings of the 30th ACM SIGKDD Conference on Knowledge Discovery and Data Mining}, 2024, pp. 5430--5441.

\bibitem{xia2023sheared}
M.~Xia, T.~Gao, Z.~Zeng, and D.~Chen, ``Sheared llama: Accelerating language model pre-training via structured pruning,'' \emph{arXiv preprint arXiv:2310.06694}, 2023.

\bibitem{xia2018dota}
G.-S. Xia, X.~Bai, J.~Ding, Z.~Zhu, S.~Belongie, J.~Luo, M.~Datcu, M.~Pelillo, and L.~Zhang, ``Dota: A large-scale dataset for object detection in aerial images,'' in \emph{Proceedings of the IEEE conference on computer vision and pattern recognition}, 2018, pp. 3974--3983.

\bibitem{cheng2022anchor}
G.~Cheng, J.~Wang, K.~Li, X.~Xie, C.~Lang, Y.~Yao, and J.~Han, ``Anchor-free oriented proposal generator for object detection,'' \emph{IEEE Transactions on Geoscience and Remote Sensing}, vol.~60, pp. 1--11, 2022.

\bibitem{sun2022fair1m}
X.~Sun, P.~Wang, Z.~Yan, F.~Xu, R.~Wang, W.~Diao, J.~Chen, J.~Li, Y.~Feng, T.~Xu \emph{et~al.}, ``Fair1m: A benchmark dataset for fine-grained object recognition in high-resolution remote sensing imagery,'' \emph{ISPRS Journal of Photogrammetry and Remote Sensing}, vol. 184, pp. 116--130, 2022.

\bibitem{cheng2017remote}
G.~Cheng, J.~Han, and X.~Lu, ``Remote sensing image scene classification: Benchmark and state of the art,'' \emph{Proceedings of the IEEE}, vol. 105, no.~10, pp. 1865--1883, 2017.

\bibitem{rahnemoonfar2021floodnet}
M.~Rahnemoonfar, T.~Chowdhury, A.~Sarkar, D.~Varshney, M.~Yari, and R.~R. Murphy, ``Floodnet: A high resolution aerial imagery dataset for post flood scene understanding,'' \emph{IEEE Access}, vol.~9, pp. 89\,644--89\,654, 2021.

\bibitem{chen2020spatial}
H.~Chen and Z.~Shi, ``A spatial-temporal attention-based method and a new dataset for remote sensing image change detection,'' \emph{Remote Sensing}, vol.~12, no.~10, p. 1662, 2020.

\bibitem{zhang2023spatial}
Z.~Zhang, L.~Jiao, L.~Li, X.~Liu, P.~Chen, F.~Liu, Y.~Li, and Z.~Guo, ``A spatial hierarchical reasoning network for remote sensing visual question answering,'' \emph{IEEE Transactions on Geoscience and Remote Sensing}, vol.~61, pp. 1--15, 2023.

\bibitem{jin2021temporal}
P.~Jin, L.~Mou, Y.~Hua, G.-S. Xia, and X.~X. Zhu, ``Temporal relations matter: A two-pathway network for aerial video recognition,'' in \emph{2021 IEEE International Geoscience and Remote Sensing Symposium IGARSS}.\hskip 1em plus 0.5em minus 0.4em\relax IEEE, 2021, pp. 8221--8224.

\end{thebibliography}

\end{document}